\documentclass[11pt]{article}

\usepackage[utf8]{inputenc}
\usepackage[T1]{fontenc}
\usepackage{lmodern}
\usepackage[margin=1in]{geometry}
\usepackage{graphicx}
\usepackage{booktabs}
\usepackage{amsmath,amssymb}
\usepackage{microtype}
\SetProtrusion{encoding = T1, family = lmtt}{ - = {0,250} }
\usepackage{caption}
\usepackage{fancybox}
\usepackage{booktabs}
\usepackage{authblk}
\usepackage{xcolor}
\usepackage[hidelinks]{hyperref}

\graphicspath{{./}{../figures/}{figures/}}

\title{\textbf{Magnolia: A Persistent-Memory Agent Framework\\ for Computational Chemistry}}
\author[1,*]{Tao Jiang}
\author[1,2]{Benjamin Navet}
\author[1,2]{Yousouf Taghzouti}
\author[2]{Fabien Gandon}
\author[1,*]{Louis-F\'elix Nothias}
\author[1,*]{Elise Dumont}
\affil[1]{Universit\'e C\^ote d'Azur, CNRS, ICN, Nice, France}
\affil[2]{Inria, Universit\'e C\^ote d'Azur, CNRS, I3S, France}
\affil[*]{Corresponding authors: tao.jiang@univ-cotedazur.fr, elise.dumont@univ-cotedazur.fr, louis-felix.nothias@univ-cotedazur.fr}

\date{\today}

\begin{document}
\maketitle

\begin{abstract}
Chemistry has recently entered a high-data regime as molecular databases from high-throughput platforms or computational studies have become available. Increasing automation within the chemical sciences leads to a more digital and precision chemistry. Dedicated scientific agents are becoming central to guiding the chemist through an ever-increasing amount of data and findings, including negative ones.
Here, we introduce Magnolia, a persistent-memory artificial intelligence agent framework that integrates computational protocols, from docking to quantum chemistry, lab notebooks and chat logs relying on large language models (LLMs). It relies on a three-tier memory system: session-level observations are automatically captured and distilled into structured learnings, confirmed and elevated to project-tier entries and, when human-validated, promoted to cross-project rules that actively guide subsequent work and provide direct chemical feedback.
We illustrate Magnolia on four case studies: quantum-chemical evaluation of aromatic systems (COMPAS database), molecular dynamics of butane and of modified DNA, and a protein--peptide docking campaign.
Across these projects, Magnolia tracked 317 runs and accumulated 116 project-tier entries, from which four cross-project rules were promoted; in controlled replays of four tasks, runs with memory passed the checks that runs without it failed (for example, ten of ten docking-configuration checks, against one to four of ten).
By combining automation, reproducibility, and human-in-the-loop control, our Magnolia framework streamlines docking, molecular dynamics and quantum chemistry calculations with direct analysis and LLM-augmented feedback, offering accelerated access to computational (bio)chemistry.

\end{abstract}

\section{Introduction}

Computational (bio)chemistry has reached a stage of maturity as projects now typically combine several methods from docking to all-atom molecular dynamics (MD) simulations and quantum chemistry~\cite{sabe2021}. Quantum-chemical datasets and databases provide a growing resource for this work \cite{Ullah-chemrxiv2024}. High-performance computing (HPC) resources allow intensive studies with several hundred quantum-mechanical (QM) calculations, thousands of docking calculations and an interplay with all-atom MD simulations up to microsecond-range trajectories. This abundance of increasingly accurate data naturally calls for robust analysis protocols and careful visual inspection of structures and MD trajectories.

Yet in virtual screening, the field's highest-throughput engine, analytical practice typically stops at score-based triage, with visual inspection of top-ranked poses still a crucial step \cite{fischer2021}. Systematic contact-level analysis that would explain why a campaign succeeded or failed remains rare: interaction-fingerprint methods serve mainly as rescoring devices \cite{marcou2007}, although tools such as ProLIF support explanatory analysis \cite{prolif2021}, and the per-residue decomposition needed to rationalize rankings is deferred to later project stages or never performed. Such depth requires retained, comparable run histories, precisely what current workflows discard. It also requires analysis capability: writing custom scripts to mine each project's outputs is prohibitive for many researchers.

To take full advantage of ever-increasing experimental or computational data, which are produced in the labs routinely, one needs new computational strategies or workflows to better capture the chemical lessons from intensive computational studies. Protocol insights, custom scripts and error resolutions are often scattered across lab notebooks, local directories and execution logs, with limited standardization or systematic reporting. These records are not always shared in full, which complicates the reproduction of a study~\cite{encore2024}. Negative findings, often valuable for preventing wasted effort, are systematically lost because publication venues rarely reward reporting them \cite{fanelli2012}. When a new graduate student joins a project or a lab moves to a new protein target, the accumulated wisdom of prior work is effectively invisible: it must be rediscovered through the same trial and error that produced it.

Workflow managers and electronic lab notebooks partially address these limitations. Tools like Snakemake~\cite{snakemake}, Nextflow~\cite{nextflow}, and FireWorks~\cite{fireworks} capture \emph{what} was run: the command, the parameters, the exit code. Electronic lab notebooks record \emph{what} was obtained or observed: operating conditions, yields, spectra, plots; they can also preserve procedures and reasoning. Executable notebooks, such as those edited in JupyterLab~\cite{kluyver2016}, combine the code, narrative and outputs of an analysis. Git~\cite{progit2014} records changes in scripts and protocols, making revisions traceable and reversible, but it versions the artifacts, not the lessons extracted from them.

Large language model (LLM) agents have recently been applied to scientific workflows, including computational (bio)chemistry, demonstrating that natural language can drive tool invocation, result parsing, and error handling \cite{chemgraph,dynamate,cascadel,hu2026aitomia,legrand2026mimosa}; a perspective surveying the field counts forty-nine such systems for atomistic simulation as of August 2026 \cite{dral2026agents}. Most of these systems, however, treat each session as an independent event; CASCADE's consolidated memory and Mimosa's workflow archive are the exceptions. The agent that helped refine a protocol on Friday has no memory of what it learned when work resumes on Monday.

This gap is acknowledged by the field itself: the developers of one state-of-the-art molecular-dynamics agent note that they ``do not include persistent memory for the agent, so each new run has no recollection of any prior runs'' and flag long-term memory as future work \cite{dynamate}; post-mortems of fully autonomous research attempts identify ``memory and context degradation across long-horizon tasks'' among recurring failure modes, with models that ``systematically lost track of previous decisions'' as sessions progressed \cite{whynotscientists}; long-horizon autoresearch frameworks still ``lack mechanisms for continuity, recovery from dead ends'' \cite{scienceflow}; and surveys of agent memory, from early context-management systems such as MemGPT to recent position papers, conclude that agents require ``proactive memory management rather than ephemeral, forgetful behaviors'' \cite{memgpt,memsurvey2025}. The atomistic-simulation agent survey classifies architecture, autonomy and delegation, but does not systematically assess persistent cross-session memory \cite{dral2026agents}. A further challenge is to distinguish what the machine learns from what a human has confirmed. The aim is therefore a system that \emph{accumulates} protocol knowledge across sessions and projects, compressing experience into durable rules that actively shape future behavior.

Such a system sits at a boundary between autonomous execution and supervised assistance that the agent community has only begun to name: as argued for skill-acquiring agents, ``tool use is not a unique ability of humans; what distinguishes humans is the cumulative acquisition of skills'' \cite{cascadel}. At one extreme, attempts at fully autonomous research have shown reliability limitations: one case study reported implementation or evaluation failure in three of four attempts \cite{whynotscientists}. Agent Laboratory also reported higher mean scores from automated reviews than from human reviewers (6.1 versus 3.8 out of 10) \cite{agentlaboratory}. On the other hand, the CASCADE authors identify sustained skill mastery, human--agent collaboration and memory-based consolidation as areas that remain underexplored \cite{cascadel}. Between these extremes lies a middle ground: persistent memory with human-gated promotion, in which the agent organizes and surfaces evidence (contact maps, scoring tables, reproducibility metrics) while the researcher retains interpretive authority. Persistent memory addresses the first barrier, agent-generated analysis the second, with the aim of making contact-level explanation more accessible. This posture mirrors the grounding discipline of deployed industrial research assistants, whose responses are ``grounded in retrieved evidence and linked back to the original sources'' \cite{astrazenecara}, and aligns with broader evidence from human--AI research collaboration, where ``every step required rigorous human verification'' \cite{airesearchpoc} and staged human feedback improved several assessed aspects of output quality \cite{agentlaboratory}.

The design of such a memory can draw on cognitive science. Long-lived agents need more than a long context window: episodic memory stores instance-specific experience rapidly, and a complementary consolidation process distills it into durable, general knowledge \cite{episodicmemory2025}. This distinction between fast, specific experience and slow, consolidated knowledge suggests a tiered memory architecture.

We introduce \textsc{Magnolia}, a persistent-memory artificial intelligence (AI) agent framework for computational chemistry that realizes this middle ground through a three-tier memory architecture:
\begin{enumerate}
\item Tier 1, the \emph{session tier}, automatically captures tool invocations, exit codes, and error traces, then distills the conversational context into structured learnings via LLM-driven extraction.
\item Tier 2, the \emph{project tier}, holds confirmed entries (error resolutions, success patterns, failure modes, parameter guidance) that are promoted from staging by human confirmation or, for entries recurring across sessions at high confidence, automatically.
\item Tier 3, the \emph{reviewed tier}, elevates project-level learnings into durable cross-project knowledge, routed by content class: operational rules are loaded into the agent's context at every session start and govern subsequent behavior, while task-shaped protocols are retrieved as on-demand skills. Rules can encode anything from protocol checklists (validate all input PDBs before submission) to tool-specific pitfalls (template-built peptide geometry can yield unphysical bond lengths) to scoring conventions (use AIR-free scores, not raw HADDOCK scores, for cross-variant comparison).
\end{enumerate}

Promotion from the project tier to the reviewed tier is human-gated by design: an LLM proposes candidate elevations based on observation frequency and consensus panel judgment, but only a human confirms them. Magnolia's role throughout is evidential rather than prescriptive: it amplifies the researcher's ability to learn from their own data; it does not replace their judgment. This architecture directly targets two recurring losses in computational chemistry workflows: undiagnosed systematic protocol biases and the silent discard of negative results. Its memory architecture offers a basis for adaptation to other computational science domains.

\section{Methods}

\subsection{Overview}

\begin{figure}[htbp]
\centering
\includegraphics[width=0.72\linewidth]{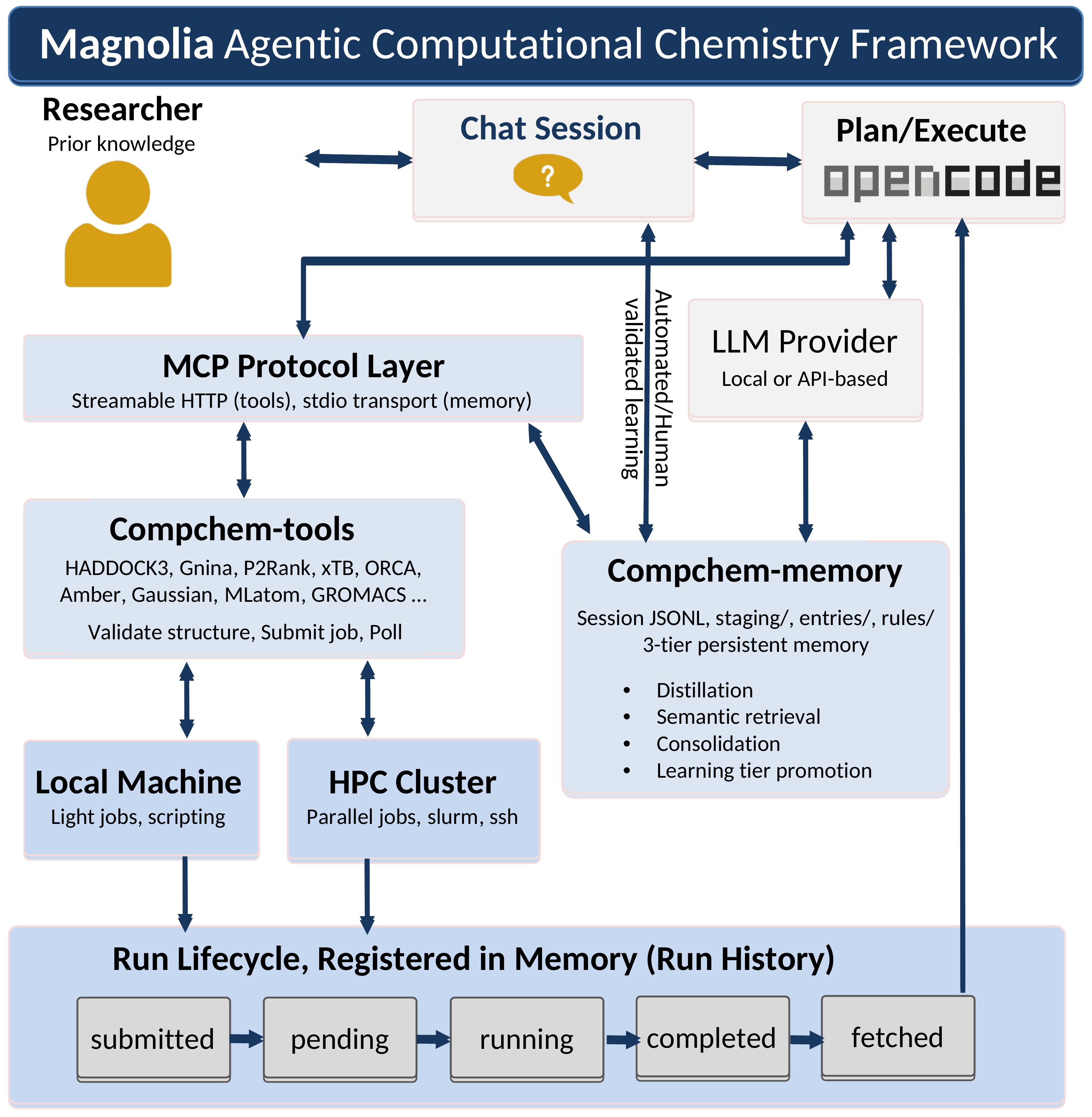}
\caption{System architecture. The researcher interacts in natural language with the OpenCode agent, which communicates with two MCP servers (\texttt{compchem-tools} and \texttt{compchem-memory}), using stdio for memory and streamable HTTP for the tool service. Tools dispatch jobs to the local machine (light jobs) or to a remote HPC cluster (\texttt{ssh-slurm}); every run is registered in the memory run history. The LLM provider, local or API-based, serves both agent reasoning and memory distillation.}
\label{fig:arch}
\end{figure}

Magnolia combines persistent project memory, tool orchestration, and HPC job management. The current implementation uses OpenCode~\cite{opencode} as its \emph{agent harness}, the runtime that manages the conversation, model access, instructions, and tool calls. Two Python services expose memory and execution operations through the Model Context Protocol (MCP). An application programming interface (API) defines how software operations are accessed; MCP standardizes how an agent discovers and invokes exposed tools. The tool handlers in turn call Python functions and scientific executables, while model-provider APIs supply model responses. Direct API or command-line integration is possible in principle, but would require an adapter that preserves capture, retrieval, and review. The persistent-memory workflow is therefore conceptually separable from its current interface. Adapting it to another harness would also require adapters for session identity, transcript export, tool-result parsing, instruction loading, and the retrieval and review hooks; such compatibility has not been evaluated here. Figure~\ref{fig:arch} shows the current implementation.

\subsection{Memory Architecture}

\begin{figure}[htbp]
\centering
\includegraphics[width=0.9\linewidth]{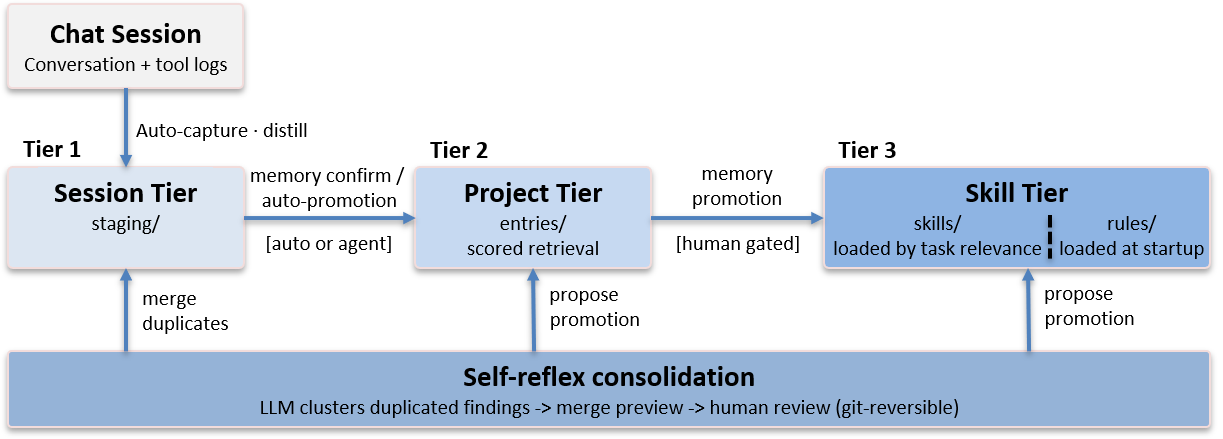}
\caption{Memory architecture. The three-tier promotion ladder: session (\texttt{staging/}) $\rightarrow$ project (\texttt{entries/}) $\rightarrow$ reviewed (operational rules in \texttt{rules/} load at every session start; task-shaped skills in \texttt{skills/} are retrieved by task relevance), fed by chat-session distillation, with an implemented self-reflex consolidation loop that merges duplicate findings (human-reviewed, Git-reversible); rule-elevation proposals are generated automatically but applied only on human confirmation.}
\label{fig:mem}
\end{figure}

The memory system is organized in three tiers with increasing durability and scope (Figure~\ref{fig:mem}).

Memory is deliberately simple: plain Markdown files with YAML frontmatter, versioned with Git. Session logs, provisional staging entries, and confirmed project entries live in the project's \texttt{.magnolia/} directory, and the reviewed rules live alongside the code; a Python MCP server manages the store, and each entry carries structured frontmatter (type, tags, confidence, observation count, verification date, provenance). Rather than a database or an embedding index, files were chosen to keep the researcher as the final reviewer: at the scale of a research project (hundreds of entries, not millions) every learning must remain human-readable, greppable, and diffable, revision history comes for free, and the agent reads the store natively. The tier structure evolved from practice (the flat session log was the first memory; the tiers were added as noise accumulated), not from a benchmarked comparison. Retrieval is heuristic scoring over the entry metadata with an optional LLM re-ranking pass; semantic retrieval was deliberately deferred at the design stage until keyword retrieval becomes the limiting factor. Mapping the entry schema to a memory ontology is future work.

\textbf{Tier 1: Session tier.} Every tool invocation (shell commands, Slurm submissions, result parsers) is logged to a session JSONL file via the \texttt{magnolia-run} wrapper. Log entries record the command, working directory, exit code, stdout, stderr, and wall time. In parallel, the full conversational transcript is exported from OpenCode and processed by an LLM-driven distillation pass that extracts structured learnings: error resolutions (symptoms $\rightarrow$ cause $\rightarrow$ fix), success patterns (protocol $\rightarrow$ quantitative result $\rightarrow$ caveat), failure patterns (attempted approach $\rightarrow$ failure mode), and parameter guidance (specific value $\rightarrow$ context $\rightarrow$ justification). Both the tool-log extraction and the conversation distillation write to a shared \texttt{staging/} directory as Markdown files with YAML frontmatter. A confidence score and observation count are assigned to each entry. Each score is the distilling model's own certainty estimate, assigned from a fixed rubric: first-observation patterns score 0.5--0.7, and 0.8 or above requires strong within-session evidence (multiple confirmations, clean metrics, or explicit user confirmation); entries without a supplied score default to 0.5. Scores are a triage signal, not a calibrated probability, and are adjusted by use: a failed run decrements the scores of success patterns that name the failing tool, and merging a duplicate keeps the higher score.

\textbf{Tier 2: Project tier.} Staging entries are confirmed via the \texttt{memory\_confirm} tool (a human review gate, not an automatic classifier), which moves them to \texttt{entries/} within the project's \texttt{.magnolia/} directory. Confirmed entries carry provenance metadata (originating session ID, timestamp, source path) and are indexed for retrieval. The \texttt{memory\_get\_context} tool is a budgeted retrieval method using heuristic scoring (title, description, tags, tools, entry type, confidence, observation count, a recency decay, and recent run outcomes) to select relevant entries for each new task, with an optional LLM re-ranking pass over the top candidates, assembling a ranked context within a configurable token budget. Entries can also be created directly by the human researcher via \texttt{memory\_annotate} (a manual authoring operation that bypasses distillation). Near-duplicate learnings are merged into the existing staging entry at write time. Entries are admitted to the project tier automatically only when they recur in at least two distinct sessions at confidence at least 0.85 and a deterministic check finds no parameter-level conflict with an existing entry (same parameter, different recorded value); flagged candidates are held for review, and the remainder stays in staging until the researcher confirms it. Admission therefore combines independent recurrence, the confidence threshold, and a clean conflict check, rather than the score alone.

\textbf{Tier 3: Reviewed tier.} Project-tier entries that have been observed across multiple sessions can be elevated into durable, cross-project knowledge. Elevated knowledge is routed by content class: behavioral gates and always-on checks become rules (\texttt{rules/}), plain Markdown files loaded into every agent's context at session start on all projects and shaping subsequent behavior; the boot inventory currently covers scheduler semantics, cluster access procedures, job-execution conventions, pre-job validation, and general agent behavior. For example, the \texttt{prejob\_check.md} rule prescribes a mandatory verification checklist before every \texttt{submit\_job} call (the verification tools exist, but invoking them remains agent-initiated rather than hard-wired into \texttt{submit\_job}). Task-shaped protocols (for example, the HADDOCK3 skill consolidating actpass formatting requirements, scoring-weight conventions, and common error resolutions) become on-demand skill documents (\texttt{.opencode/skills/}), loaded only when a task matches; a protocol costs no context in other sessions. Promotion is human-gated: the proposal pipeline described below drafts rule candidates automatically, and the researcher decides acceptance and, for task-shaped knowledge, its placement in the appropriate skill. Rule and skill files are versioned with a \texttt{last\_verified} timestamp to support staleness detection.

One transformation illustrates the path. During the protein--peptide docking campaign (Case Study 4), three misreports of AIR-free scores (subtracting the full restraint term, using the wrong restraint weight, and transcription slips) accumulated into a single project entry carrying the explicit formula and its sanity checks. That entry was distilled into the AIR-free section of the HADDOCK3 skill, which cites the entry as its evidence base; the entry remains as the record of the incidents, and the checks are now part of the skill.

\begin{quote}
\small
\texttt{name: air-free-score-calculation-error (abridged)}\\
\texttt{type: error\_resolution; confidence: 0.9; observation\_count: 3}\\
\texttt{tags: HADDOCK3, AIR-free-calculation, error-correction}\\[3pt]
During presentation of comparative docking scores, the AIR-free value for a 6-mer
variant was incorrectly reported as $-157.3$ instead of the correct $-128.08$
(score $-124.84 - 0.1\times32.44$). The mistake was caused by subtracting the full
AIR term instead of applying the weight $w_{\mathrm{air}} = 0.1$. The fix recorded
the explicit formula and its sanity checks with the HADDOCK3 protocol rules.\\[3pt]
\textit{Derived skill, AIR-free section:} ``Read \texttt{w\_air} from
\texttt{weights\_params.json}; do not assume a default value'' (evidence line:
this entry; three misreport incidents).
\end{quote}

\textbf{Consolidation and self-reflex.} A self-reflex consolidation pass is implemented for the merge operation. An LLM clusters same-claim findings in \texttt{staging/} (detection with intelligence); deterministic code then assembles a merged preview from the source text only (it authors no new content) and writes it to a non-mutating proposal artifact (\texttt{reflex/consolidation-proposal.json}). The agent renders the pending proposals to a human-readable review file (\texttt{magnolia-review/proposals.md}); once the human accepts or rejects each one, deterministic code applies the accepted merges and commits to a nested Git repository tracking \texttt{entries/} and \texttt{staging/}, providing reversibility and an audit trail. Rejections are recorded durably (keyed by source content) so they do not re-surface on later sweeps. Promoted rules carry a \texttt{last\_verified} date and are re-verified manually: no rule expires automatically, and a rule tied to a superseded tool version stays until the researcher revises or retires it, with the old text preserved in the version history. At the project tier, a health-check tool flags stale entries (default: ninety days), low-confidence, orphaned, duplicate, or broken-reference items, and stalled staging entries; its default mode is report-only, with an optional pass that removes broken references and marks stale entries. A consolidation pass merges duplicate findings and archives the lowest-confidence entries beyond a size cap. Automatic re-verification, rule expiry, and demotion (returning a rule to the project tier) are not yet implemented. The \texttt{propose-promotion} operation, which surfaces project-tier entries ready for elevation to the skill tier, is implemented as a proposal-only pipeline: a deterministic eligibility gate (entries observed in at least three distinct sessions), a three-pass LLM consensus panel with a correctness veto, rule drafting with a faithful-copy fallback, and a consistency check against existing rules. Surviving proposals are written to a review artifact and applied only on human confirmation; rejected candidates are memorized so repeated sweeps do not re-run the model on them. Conflicts have no automatic precedence or detection: the layers have distinct roles (always-on rules shape behavior, retrieved entries supply evidence), so a rule and an entry that disagree are reconciled by the researcher. At promotion time, a drafted rule is checked against the existing rules, and duplicate or conflicting matches are surfaced to the reviewer, never auto-dropped. Rules and per-project memory stores are version-controlled, so revisions and retirements are revertible. Guards against reinforcing bad practice are the eligibility gate (at least three distinct sessions), the panel's correctness veto, and the caveats required on captured learnings; automatically detecting that a promoted rule is steering work wrongly remains future work.

\subsection{Tool Orchestration}

Scientific software packages are exposed to the agent as MCP tools, each with a typed parameter schema and structured, machine-parseable results. The non-exhaustive MCP tool inventory includes HADDOCK3~\cite{haddock3} (protein--protein and protein--ligand docking), Gnina~\cite{gnina} (CNN-scored molecular docking), P2Rank~\cite{p2rank} (ML-based pocket prediction), xTB~\cite{banwarth-gfn2-xtb} (semi-empirical quantum chemistry), ORCA~\cite{ORCA6}, MLatom~\cite{MLatom233}, Gaussian~\cite{gaussian16} (DFT quantum chemistry), and Amber~\cite{amber26} and Gromacs~\cite{gromacs2015} for molecular dynamics including Ambertools for non-canonical ligand parameterization. Each wrapper checks process exit codes and returns structured results; input-file validation is exposed as a dedicated \texttt{validate\_structure} tool plus named \texttt{stage\_gate} checkpoints (e.g., \texttt{docking\_inputs\_ready}, \texttt{scf\_converged}) that verify intermediate outputs before proceeding to downstream steps, and shell-level execution is logged through \texttt{magnolia-run}. Literature queries are served through a retrieval-augmented search interface (Perspicacité \cite{pradi2025perspicacite}) over curated chemistry knowledge bases, bringing published evidence into the same session context as computational results.

The underlying software itself is installed in isolated environments under \texttt{softwares/}, registered in a software manifest, and invoked through wrapper scripts that handle environment activation, library path configuration, and binary discovery. MCP integration is reserved for the software the agent drives routinely, where typed interfaces, automatic result parsing, and run tracking pay off; packages used more occasionally are invoked through their command-line wrappers, registered in the same manifest. BoltzGen (AI protein design), for example, is installed as a versioned venv and driven through its wrapper script. The agent discovers the available software through this manifest and probes installed versions via \texttt{check\_environment}.

\subsection{Pre-Job Validation}\label{subsec:prejob}

Before every \texttt{submit\_job} call, the \texttt{prejob\_check} rule prescribes a mandatory verification sequence (agent-initiated, not hard-wired into \texttt{submit\_job}): all input files are confirmed present and non-empty, every structure file is validated via \texttt{validate\_structure}, symlinks to external paths are rejected, and tool-specific checks are applied. For HADDOCK3 runs, these include PDB chain and segment ID verification, residue range consistency between the configuration and PDB contents, and actpass restraint file validation. The rule distills recurring configuration-mismatch errors from the early docking campaigns, the clearest being a 10-mer batch launched with 5-mer restraints that completed without errors but produced misleadingly poor scores: the failure class this checklist targets is invisible at submission time and only surfaces in the results.

\subsection{Job Execution and HPC Integration}

Computations are dispatched through a single \texttt{submit\_job} interface whose \texttt{scheduler} parameter selects the execution target. Short, light jobs run on the local machine (\texttt{scheduler="local"}): the command is launched as a background process, registered in the project run history with lifecycle \texttt{running}, and its completion is checked on demand from the process state and output files: there is no submission queue or result transfer. Long-running computations ($>$30 minutes) are routed to a remote HPC cluster via the \texttt{submit\_job(scheduler="ssh-slurm")} path. The tool establishes an SSH connection through an OpenConnect SOCKS5 tunnel to the cluster, writes an sbatch script with resource specifications (partition, cores, memory, wall time), rsyncs the run directory to cluster scratch (\texttt{/workspace}), and submits the job. A local YAML record tracks the remote run lifecycle through the states \texttt{writeahead} $\rightarrow$ \texttt{submitted} $\rightarrow$ \texttt{running} $\rightarrow$ \texttt{completed} $\rightarrow$ \texttt{fetched}. A background poller sweeps active ssh-slurm runs every 5 minutes, queries \texttt{sacct} for terminal states, auto-fetches results on completion, and triggers \texttt{post\_run\_assess} for automated metric extraction; local runs are registered in the same run history but are advanced on demand rather than by the poller. Infrastructure failures (node failures, boot failures, preemption) are distinguished from application failures; bounded automatic retry for the former is implemented on the development branch. The job layer can also resume a run in place: a restarted submission reuses the run's preserved directory and record, and supported tools continue from intermediate state (for example, a HADDOCK3 restart or a GROMACS checkpoint). Failures that exit successfully but produce wrong results are not automatically detectable: deciding what to verify is part of the scientific knowledge being accumulated. Such failures are caught by researcher review against evidence: pre-job validation for known configuration pitfalls (Section~\ref{subsec:prejob}), and cross-run and literature comparison for scientific ones (Case Study 4). A class a researcher identifies can then be encoded as an explicit check that later runs execute. The \texttt{cancel\_job} tool provides on-demand cancellation.

\subsection{Implementation}

The memory system is implemented as a Python MCP server (\texttt{compchem-memory}) with modules for entry management, heuristic retrieval with optional LLM re-ranking, consolidation, session distillation, and run history tracking. The tool orchestration layer is a separate MCP server (\texttt{compchem-tools}) with MCP wrappers for the scientific tools the agent drives routinely. The two servers communicate with the OpenCode agent through the MCP protocol, over stdio for the memory server and streamable HTTP for the tool server. The agent's behavior is governed by a collection of Markdown rule files (\texttt{rules/}) loaded at session start, covering HPC conventions (remote cluster access, Slurm semantics), job-execution procedures, and operational checklists; tool-specific protocols (HADDOCK3, GROMACS, P2Rank, xTB, Gnina, Orca, MLAtom) reside in on-demand skill documents and are retrieved by task relevance. The framework is configured as an OpenCode workspace; project scaffolding, venv activation, and MCP server startup are handled by the \texttt{magnolia} launch wrapper.

\section{Results}

In this section, we present four case studies that illustrate different task types in computational chemistry: i) a quantum-chemistry benchmark on a database subset; ii) classical versus machine-learned molecular dynamics on a textbook organic molecule; iii) the automated setup of DNA simulations; and finally iv) a six-week protein--peptide docking campaign, presented here with identifying details withheld while the collaboration is ongoing. All four ran with the same agent, each with its own project store. The first three are short campaigns that do not isolate the memory's effect; the fourth is a full-length project in which accumulated memory demonstrably shapes the work. Section~\ref{sec:memanalysis} examines what these campaigns deposited in the memory store in aggregate: the composition of the project-tier store, the evolution of its learning-type mix, and the rules elevated from it. Throughout, the cases document what the agent presented and what the researcher decided.

\subsection{Case Study 1: Systematic screening of aromatic compounds electronic properties}

We used Magnolia to perform a series of quantum chemistry calculations on all the aromatic compounds featuring up to seven rings. The latter were extracted from the COMPAS database (Computational Database of Polycyclic Aromatic Systems \cite{compas}).  179 compounds were extracted, and we verified that this number indeed matches the COMPAS database. HOMO-LUMO gaps, total energies, total dipoles and singlet-singlet vertical transition energies were computed using the semi-empirical GFN2-xTB method \cite{banwarth-gfn2-xtb}, and compared to the original COMPAS data at the B3LYP-D3BJ/def2-SVP level of theory, generated with ORCA 4.2.0 \cite{compas}. Magnolia performed a first series of QM calculations and situated the relative performance of the semi-empirical GFN2-xTB as:

{\small
\noindent\textbf{Key findings for the 179 PAHs with $n_{\mathrm{rings}} \leq 7$:}
\begin{itemize}
\item GFN2-xTB reproduces the B3LYP gap ranking nearly perfectly ($R^2 = 0.99$).
\item There is a systematic $\sim$1.2 eV gap underestimation: xTB gaps are consistently narrower. This is expected for a semi-empirical method versus hybrid DFT.
\item The bias is consistent across ring counts (1--7 rings), not dependent on system size.
\end{itemize}
}

We let Magnolia determine the screening protocol through its project memory store instead of fixing it in advance. The initial structure-generation route (direct OpenBabel 3D generation feeding GFN2-xTB, validated by 25/25 clean optimizations) was recorded as an operational note, then a scoped parameter-guidance entry claimed pre-optimization was unnecessary for rigid planar PAHs; an audit of the COMPAS-2 production code withdrew that claim in favor of an RDKit ETKDG + UFF pre-optimization sequence, captured as the reproduction protocol for future runs. The benchmark numbers reported above predate this correction, which therefore illustrates the memory system's intended maintenance loop (record, scope, revise) rather than a change to the reported data: both structure-generation routes feed the same GFN2-xTB geometry optimization, and these rigid, planar molecules carry no conformational complexity, so a rerun is not expected to change the reported gaps and IPs.

Magnolia reproduced the GFN2-xTB data reported for these 179 compounds in the COMPAS database. 
Ionization potentials (IP) under Koopmans' theorem were also compared, indicating an excellent correlation yet with a large offset which is typically encountered with semi-empirical applications. Magnolia itself then proposed fetching experimental adiabatic IPs from the NIST Chemistry WebBook~\cite{nistwebbook} to enable a comparison for a subset of nine compounds (see Table ~\ref{tab:molecules_booktabs}).

\begin{table}[h]
\centering
\caption{Experimental vs. calculated IPs (NIST Chemistry WebBook, first column) vs. computed: xTB, B3LYP and adiabatic IP with the $\Delta$SCF approach. All values are given in eV.}
\label{tab:molecules_booktabs}
\footnotesize
\setlength{\tabcolsep}{3pt}
\begin{tabular}{@{}p{4.9cm}cccc@{}}
\toprule
Molecule & IE$_{\text{exp}}$ & $-$E(HOMO) xTB & $-$E(HOMO) B3LYP & Adiabatic IP \\ 
\midrule
benzene & 9.244 & 10.95 & 6.88 & 9.00 \\
naphthalene & 8.144 & 10.27 & 5.95 & 7.73 \\
phenanthrene & 7.891 & 10.22 & 5.89 & 7.45 \\
triphenylene & 7.870 & 10.31 & 6.01 & 7.43 \\
chrysene & 7.600 & 10.04 & 5.68 & 7.09 \\
benz[a]anthracene & 7.450 & 9.91 & 5.48 & 6.90 \\
anthracene & 7.439 & 9.84 & 5.39 & 6.95 \\
tetracene & 6.970 & 9.56 & 5.02 & 6.41 \\
pentacene & 6.630 & 9.37 & 4.76 & 6.03 \\
\midrule
Level of theory & ME & MAE & RMSE & $R$ \\
GFN2-xTB $-$E(HOMO) (Koopmans) & +2.36 & 2.36 & 2.38 & 0.9887 \\
B3LYP-D3BJ/def2-SVP $-$E(HOMO) (Koopmans) & $-$2.02 & 2.02 & 2.03 & 0.9887 \\
B3LYP-D3BJ/def2-SVP adiabatic IP ($\Delta$SCF) & $-$0.47 & 0.47 & 0.48 & 0.9997 \\
\bottomrule
\end{tabular}
\end{table}

\begin{figure}[htbp]
\centering
\includegraphics[width=\linewidth]{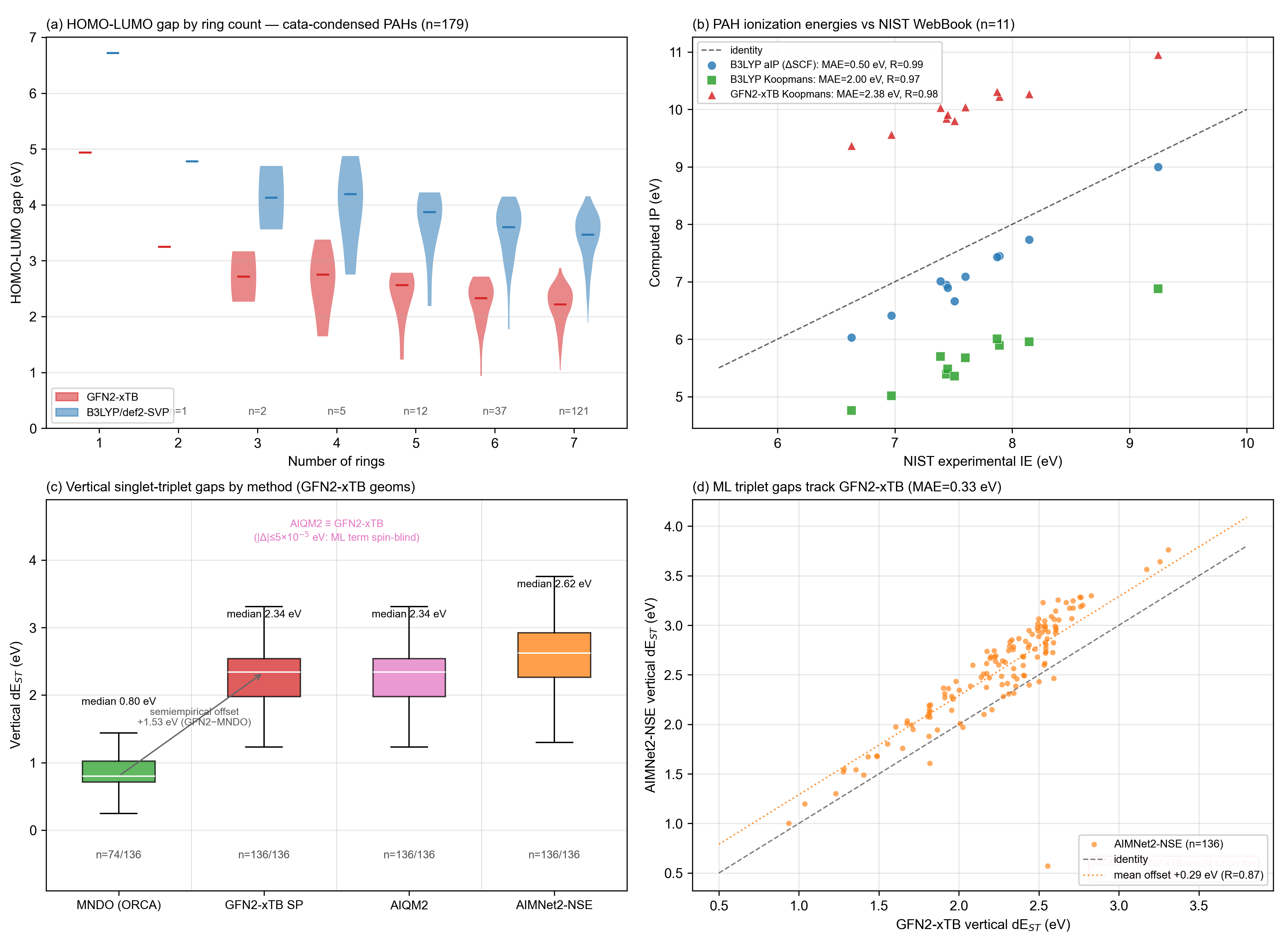}
\caption{Quantum-chemistry benchmark on the COMPAS subset. (a) HOMO--LUMO gaps by ring count for the 179 cata-condensed PAHs, GFN2-xTB versus B3LYP/def2-SVP. (b) Computed versus experimental ionization energies (NIST Chemistry WebBook); the dashed line is the identity. (c) Vertical singlet--triplet gaps by method, computed at GFN2-xTB geometries. (d) AIMNet2-NSE versus GFN2-xTB vertical triplet gaps.}
\label{fig:compas-1}
\end{figure}

We also performed excited-state calculations with the OMNI-P2x method \cite{omni-p2x} recently implemented within the MLAtom platform \cite{MLatom233} to test the operability of Magnolia with another QM suite of programs. The first ten vertical singlet-to-singlet excitations were computed. Results are gathered in Figure~\ref{fig:compas-1} and Magnolia was able to propose a systematic, accelerated exploration of electronic properties for a subset of nine aromatic compounds as listed in Table \ref{tab:molecules_booktabs}.

This first case study shows that the agent can assess the performance of QM methods over a set of molecules from a database and link it to NIST experimental data, a capability that mostly reflects the underlying LLM rather than the memory. Its memory footprint is correspondingly small: seven entries in staging (the corrected pre-optimization guidance among them, reproduced in the ESI) and no confirmations. It suggests that benchmarks piloted with Magnolia could be automated, with detailed reports shareable with the community.

\subsection{Case study 2: comparing classical and ab initio dynamics for butane}

We then move on to molecular dynamics (MD) simulations for a textbook organic molecule, butane, with the goal to situate gas-phase butane dynamics with classical MD (GAFF2 force field) vs. the AIQM2 (AI-enhanced QM Method 2, combining semi-empirical and NN \cite{aiqm2}). Setting up MD simulations is notoriously less straightforward than static QM calculations, yet the LLM was found to be particularly helpful to guide the choice of a minimization--heating--equilibration protocol either with Amber \cite{amber26} for the classical GAFF2 force field or with MLatom \cite{MLatom233} for AIQM2 MD simulations (10~ns and 50~ps respectively, running locally on a laptop).

\begin{figure}[htbp]
\centering
\includegraphics[width=\linewidth]{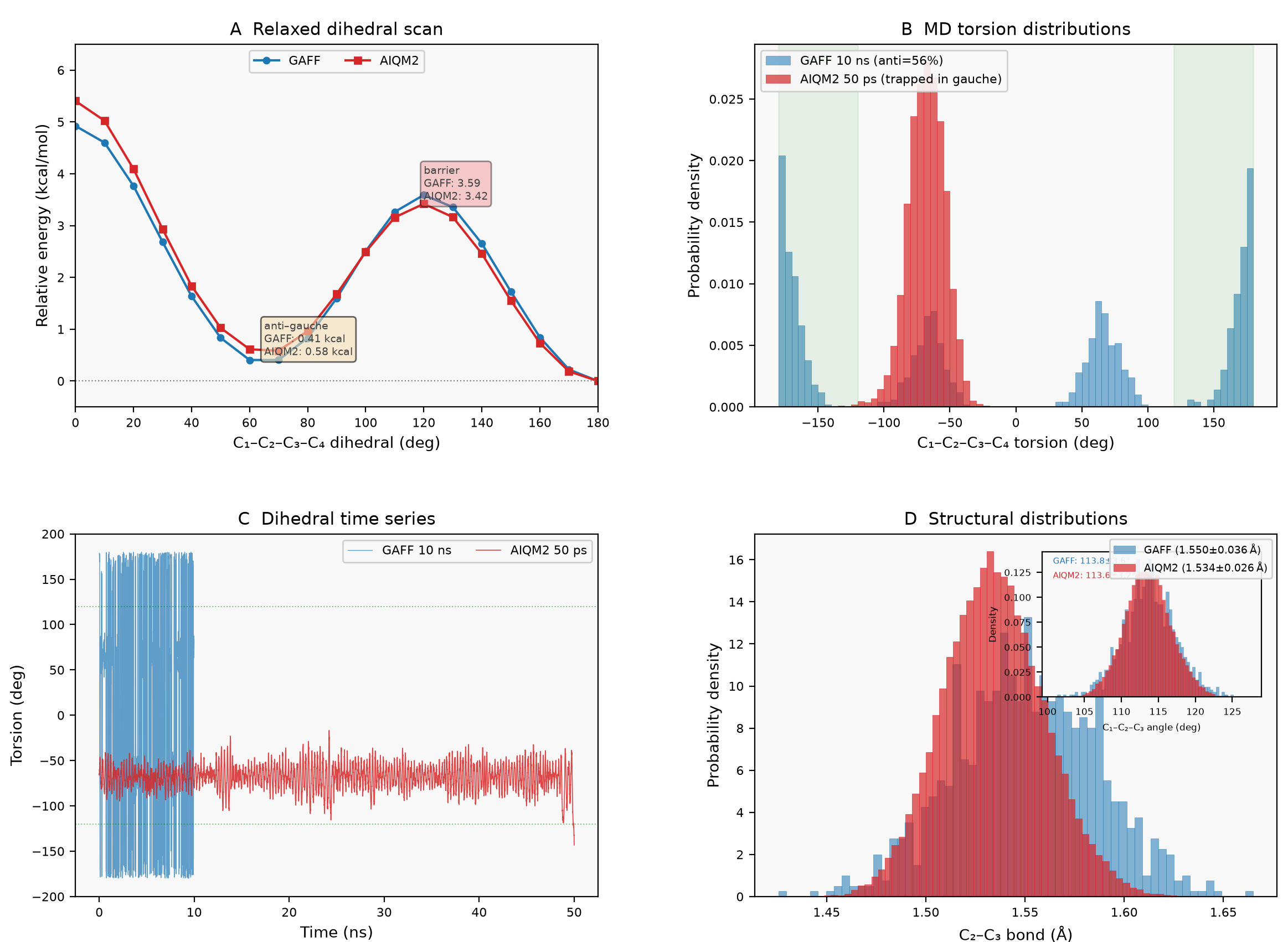}
\caption{Magnolia's automated analysis of the classical vs. AIMD molecular trajectories for butane, of respectively 10~ns and 50~ps. The AIQM2 torsion vs. time is magnified for visualization purposes.}
\label{fig:butane-analysis}
\end{figure}

Butane is a canonical, well-documented example in Organic Chemistry, to illustrate one-dimensional potential energy surfaces. The LLM conducted the post-analysis of the two trajectories to propose a relevant comparison of the conformers population respectively for GAFF2 and AIQM2 (anti/gauche population ratio, RMSD, etc. as shown in Figure \ref{fig:butane-analysis}), according to three take-home messages. Interestingly, Magnolia took the initiative to perform a relaxed scan (panel A in Figure~\ref{fig:butane-analysis}) to evaluate the energy between the different conformations.

\begin{Sbox}
\begin{minipage}{\dimexpr\linewidth-11pt\relax}
\texttt{\footnotesize{
1. 50 ps AIMD is insufficient for the butane torsion barrier (\textasciitilde3.4 kcal/mol).\\
   The trajectory never escapes the initial gauche well, so torsional
   populations are an artifact of initial conditions — not a force field
   difference. This is the central methodological finding.
\\
2. Fast DOFs are reliable even from short AIMD: C-C bond lengths and
   C-C-C angles agree well between GAFF and AIQM2.
\\
3. Relaxed dihedral scan is the correct way to compare torsional energetics
   without ns-scale dynamics. Both methods give physically reasonable results
   with the scan.
}}
\end{minipage}
\end{Sbox}
\noindent\fbox{\TheSbox}

The choice of relevant molecular descriptors is straightforward for butane with end-to-end distance C1/C4 and the corresponding, correlated dihedral angle $\tau$(C1-C2-C3-C4). For more general molecules, generative AI will be important in introducing new reaction coordinates or collective variables \cite{chipot-dd2026}. We note in passing that many MD parameters can be chosen natively by the LLM,  which calls for caution especially for larger, less canonical systems.

The memory store holds eight entries in staging from this short campaign: six operational (the AIQM1/AIQM2 platform fixes and protocol setups) and two scientific negatives (the torsion-population sampling artifact and a later conformer-search miss, each stored with the analysis that exposed it). Because usage has been short, none has yet passed the human confirmation step, and nothing has been promoted.

\subsection{Case study 3: setting up DNA MD simulations featuring a non-canonical nucleotide}

\begin{figure}[htbp]
\centering
\includegraphics[width=0.4\textwidth]{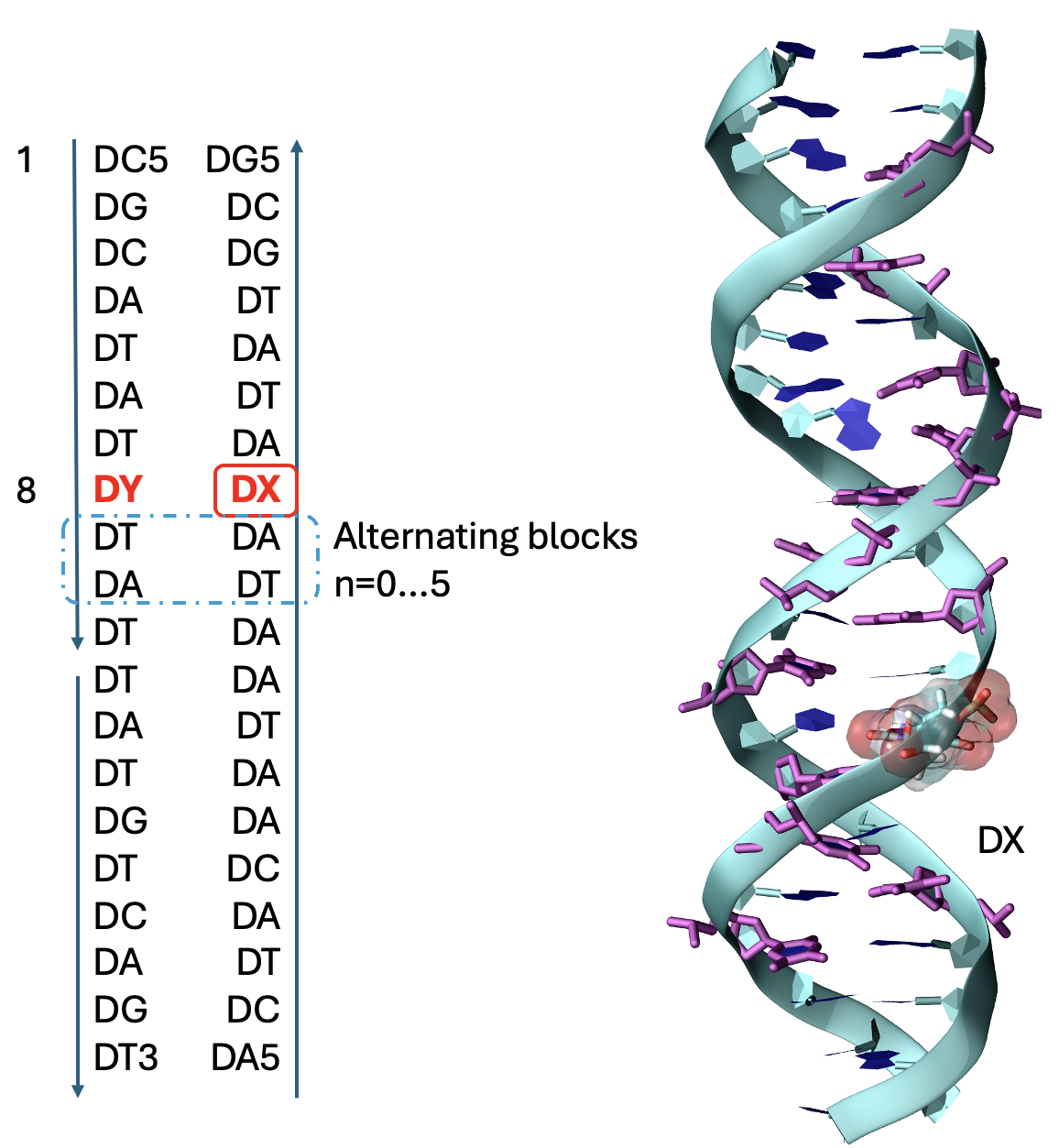}
\includegraphics[width=0.5\textwidth]{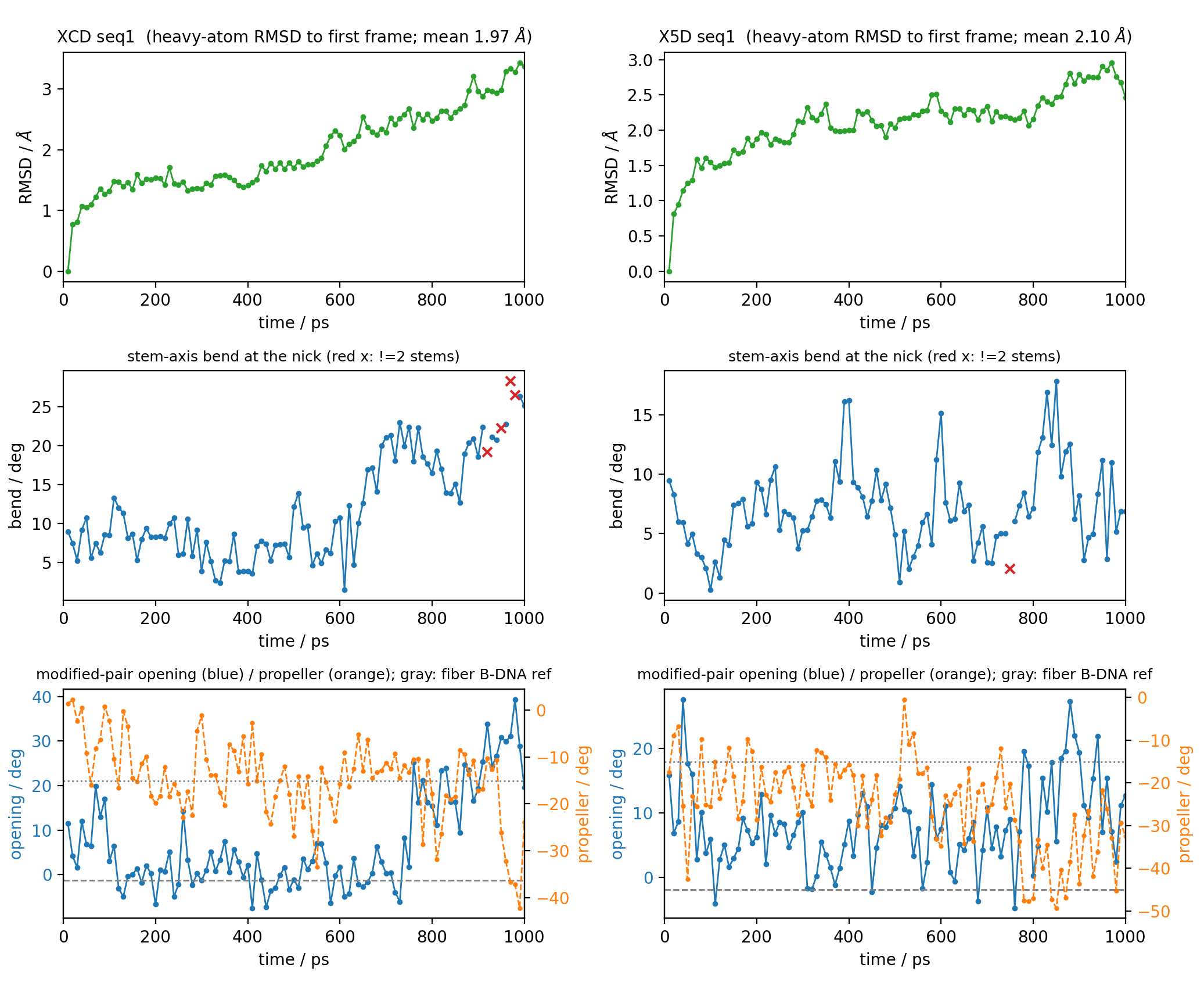}
\caption{a) Sequence of a series of six ($n$=0...5)  double-stranded sequences featuring DX=5-carboxycytosine-containing or DX=5-carboxyuracil-containing. The composition of the simulation boxes is given as ESI. b) DSSR analysis for sequences 1: RMSD, bend, opening and propeller over time.}
\label{fig:DNA-simulations}
\end{figure}

We then relied on Magnolia to automate the setup of classical MD simulations for a more complex biosystem, which requires the construction of several related oligonucleotides featuring a covalently-tethered non-canonical nucleobase (5-carboxycytosine). The structures of a series of six double-stranded DNA duplexes were generated by Magnolia and are sketched in Figure \ref{fig:DNA-simulations}. Systems were solvated in truncated octahedral OPC water boxes and the composition and MD protocol are summarized in ESI.

Magnolia installed AmberTools locally and prepared the modified DNA systems with the recent modXNA approach \cite{modxna} based on the Amber/Ambertools manual. After a two-step minimization, the system was then heated at 300 K for 100 ps, using a Langevin thermostat with $\gamma_\mathrm{ln}$ = 2 ps$^{-1}$ and equilibrated (see ESI).
 Magnolia tracked the submitted jobs on the HPC cluster through completion and retrieval, then analyzed the trajectories (energies, RMSD, local distances, temperature) to catch problems early. Like other agents (DynaMate~\cite{dynamate}, Protomia~\cite{protomia}), automation reduces the risk of human error and streamlines the daily MD workflow, especially for the same-footing comparison of DNA sequences where automation is truly needed. Full analysis including DNA helicity and associated descriptors by the 3DNA-DSSR \cite{dssr2015} were also  performed as a post-analysis to evaluate the impact of the modified nucleobase on B-helicity and provide a complete analysis of characteristic angles and distances.

The campaign also left a substantial setup-level memory footprint: seven confirmed entries and twelve in staging. They record what made the systems buildable (the modXNA/C5C setup on the local toolchain, a coordinate-injection workflow that recovers B-DNA geometry after tleap rebuilds the duplex, the root cause of a sander restart-format failure, the H-displacement fix in the injection script) and the parameters the series rests on (the modified-strand rebuild protocol and the inter-strand centre-of-mass distances across the six duplexes). None was promoted to a rule; the entries stay project-scoped. 

%
%
%
%

\subsection{Case study 4: A Protein--Peptide Docking Campaign, from Failed Restraints to Systematic Extension}

This last example, from computational biochemistry, is an ongoing study with experimental collaborators; the protein identity, peptide sequences, and structural identifiers are withheld pending the collaborators' own publication and will be released in a later version. The claims of this paper concern the framework's operation (the runs tracked, the entries captured, and how evidence was presented), none of which depend on the undisclosed details.

The protein--peptide docking campaign began on a predicted full-length receptor structure supplied by the experimental collaborators, with the binding site defined from experimentally identified active residues in a regulatory subdomain. Throughout this section, docking scores are reported as AIR-free values (the HADDOCK score minus its restraint-violation contribution), which makes runs under different restraint schemes comparable. Early HADDOCK3 runs under the literature-defined restraints failed to produce plausible complexes, with AIR-free scores ($-59$ to $-83$ across five peptides) that were the worst of the stage. The agent presented per-cluster metrics and per-residue contact maps for each run, making the pattern visible; the failure was recorded as a project-tier failure pattern rather than discarded. Switching the restraints to a predicted pocket on the same receptor then produced scores 20--45 points lower ($-85$ to $-112$) that initially looked like a solution.

Both the error and its correction traced to the literature. The original binding-site definition came from a preprint provided by the experimental collaborators. When the predicted-pocket runs overwhelmed the literature-defined site in score, the conflict, visible in the agent's cross-run comparisons, prompted a further discussion. The collaborators pointed to a revised version of the preprint, which proposed a different pocket and clarified that the relevant binding conformation is the closed, peptide-bound state of the substrate-binding domain. At the researcher's request, Magnolia then searched the literature through its Perspicacité retrieval system \cite{pradi2025perspicacite}. The revised preprint, cross-referenced with a 2.0~\AA{} crystal structure of that domain, established the correct binding groove; the crystal structure was a higher-quality template for defining the site than either the predicted full-length model or the collaborators' own predicted complexes. The same search surfaced a peptide reported as one of the tightest known ligands of this receptor family, which seeded the design work described below.
The store recorded the successive steps of this transition: the experimental concern, the agent-conducted literature finding, and the template recommendation that followed.

The crystal structure redefined the binding site rather than the working receptor. The pocket was re-derived from the 19 crystallographic residues in tight contact with the co-crystallized peptide, in place of the 8 residues proposed from the predicted models. A structural comparison recorded in the store showed that the groove backbone agreed to within 1~\AA{} per residue between the predicted and crystal structures, while the flanking architecture differed by 4--7~\AA{}: restraint definition, unlike rough groove geometry, required the experimental template. Because the groove itself is conserved, the crystal-derived residue set transferred directly onto the full-length predicted model, which remained the working receptor for the rest of the campaign. Two tool limitations were captured in the same episode: pocket prediction could not detect the shallow substrate groove, and on the full-length model it had silently top-ranked an unrelated deeper cavity. The strongly negative scores of the predicted-pocket runs were thereby revealed to reflect docking against the wrong site; one extension batch had been docked against that cavity entirely.

\begin{figure}[htbp]
\centering
\includegraphics[width=\linewidth]{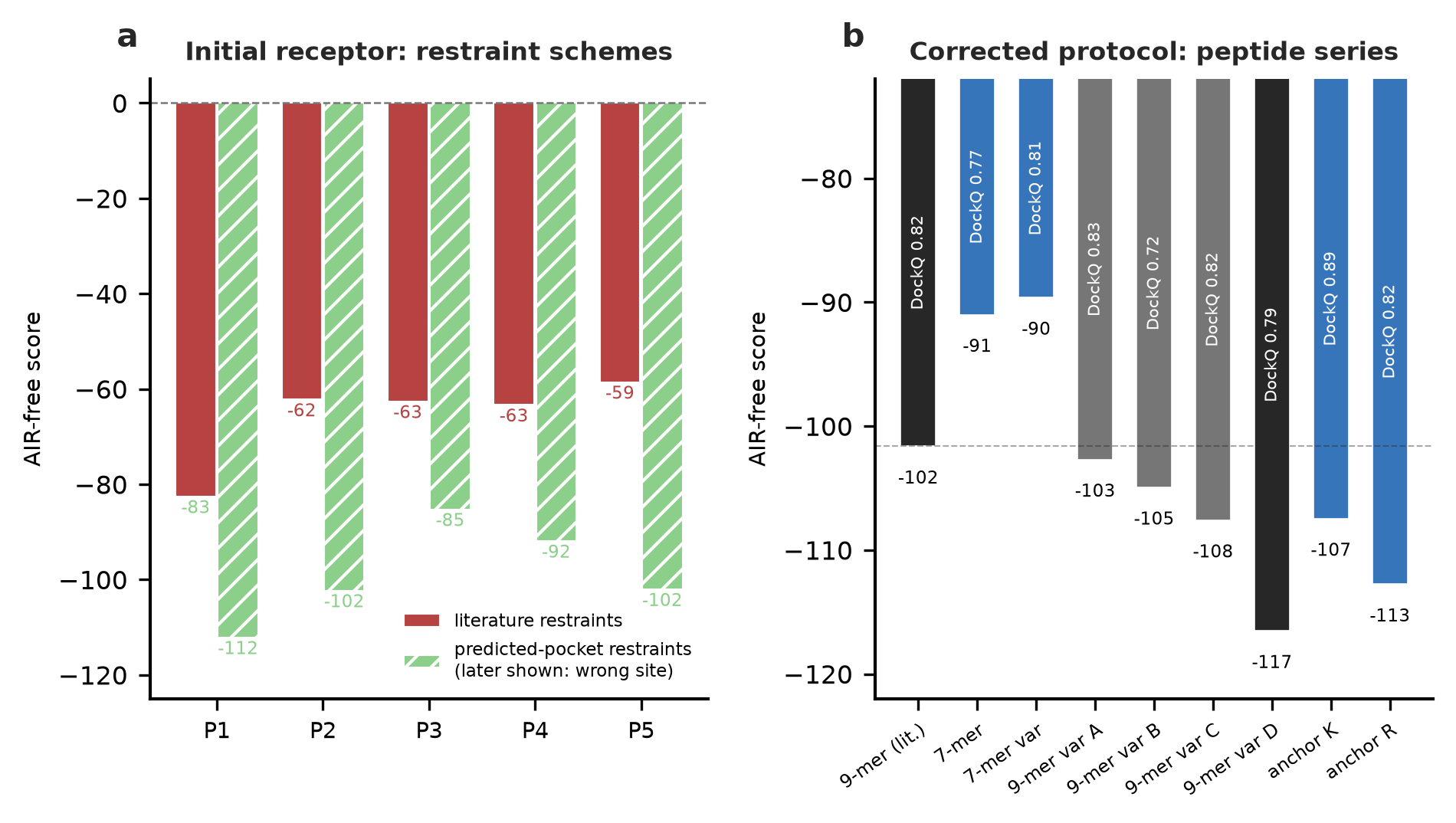}
\caption{Protein--peptide docking. (a) AIR-free scores on the initial full-length receptor under literature-defined versus predicted-pocket restraints; the apparent improvement under predicted pockets was later shown to reflect docking against the wrong site. (b) AIR-free scores on the corrected protocol (full-length receptor, crystal-contact restraints): the literature-derived 9-mer, the shortened 7-mer, and single-residue variants. Best-defined (highest-DockQ) cluster per run; peptide identities withheld; DockQ is self-referential (cluster tightness, not accuracy against a reference complex).}
\label{fig:restraint}
\end{figure}

The final protocol combined the full-length predicted receptor with restraints built from the crystal-contact residues (active and passive definitions generated via \texttt{haddock3-restraints}); no pocket-prediction step remained. On this protocol, the literature-derived 9-mer docked with its best cluster at $-101.6$ AIR-free and a well-defined binding geometry (self-referential DockQ 0.819, hereafter self-DockQ; see Figure~\ref{fig:restraint}), and the per-residue contact maps showed the hydrophobic core carrying the interface while the terminal residues made almost no contacts (Figure~\ref{fig:restraint}). Guided by this analysis and the published alanine scan, the researcher shortened the peptide to a 7-mer that retained the contacting core; it scored $-91.0$ AIR-free (self-DockQ 0.770) and became the parent of subsequent single-residue mutation scans.

Tandem and fusion constructs scored better still (down to $-137$ AIR-free), but the contact and alanine-scan analyses showed the longest spanning 43~\AA{} between the groove and a second pocket, with six of fourteen residues contributing nothing; whether such a bridge can cross-link and disable the protein remains untested, and subsequent design was capped at 9--10 residues, which contact the groove only. Within that cap, 9-mer derivatives of the literature peptide carrying a designed anchor at the groove entrance reached $-116.5$ AIR-free with well-defined geometry (self-DockQ 0.78--0.89), and two of them raised the fraction of poses in the groove-bound mode from 4.5\% to roughly half, a change in pose distribution (not a binding free energy) that the raw scores alone do not show. The researcher made each design decision; Magnolia's role was to present the contact maps and score histories that made each pattern visible.

Across its full duration the campaign produced 195 tracked runs and 109 project-tier entries across its two stores (snapshot 2026-09-02; Table~\ref{tab:stores}); Section~\ref{subsec:temporal} analyzes this store. The binding-site correction is visible there as a chain of linked entries (failure pattern, literature finding, template recommendation, tool limitation, corrected pocket definition) rather than a single event, which is the granularity at which the correction can be audited and reused.

During the campaign, blind docking with HADDOCK3's cmrest protocol produced unexpected score variability: identical peptide sequences scored differently depending on their starting conformation, with a $\sigma \approx 26$ spread across 26 runs and differences up to 30+ points in individual comparisons. The agent's side-by-side presentation of scoring tables across protocols showed the variability was systematic: scores tracked input geometry rather than sequence. Switching to a surface-derived \texttt{ambig.tbl} protocol reduced the within-sequence spread to $\sigma \approx 5$ across 15 validation runs.

\section{Memory-System Analysis}\label{sec:memanalysis}

Each case study closed with the memory footprint of one campaign. This section examines the store in aggregate: what the campaigns deposited across projects, how the composition of the largest store changed over six weeks, and what was elevated from entries into cross-project rules; it closes with a controlled replay that tests whether what accumulated changes what the agent does.

\subsection{Store Composition}

Across its four case studies, the Magnolia framework logged 317 tracked runs through its lifecycle manager (snapshot 2026-09-02; Table~\ref{tab:stores}), spanning HADDOCK3 docking (189 runs), molecular dynamics of butane and of modified DNA with Amber and MLatom (100 runs), smaller batches of BindCraft and P2Rank runs (5 runs), and 23 further records with no classified tool. The auto-extraction pipeline produced 116 project-tier entries in total (human-confirmed or auto-admitted), distributed across six learning types: 23 scientific findings, 40 general observations and corrections, 14 failure patterns (approach $\rightarrow$ failure mode $\rightarrow$ lesson), 16 error resolutions (symptoms $\rightarrow$ cause $\rightarrow$ fix), 13 parameter guidance entries (specific value $\rightarrow$ context), and 10 success patterns (protocol $\rightarrow$ result $\rightarrow$ caveat). Table~\ref{tab:inventory} provides a categorized inventory.

\begin{table}[htbp]
\centering
\caption{Project stores of the four case studies at the snapshot of 2026-09-02. Runs are lifecycle records dated up to the snapshot; staging and project-tier counts are read from the versioned store histories (index files excluded). Case Study~1's benchmark ran as shell-script batches outside the lifecycle manager; it therefore has no run records. The docking campaign comprises two stores, listed separately; the June--July store is the one analyzed in Figure~\ref{fig:entrymix}. Promoted rules are learnings elevated from entries into durable, human-gated rules.}
\label{tab:stores}
\small
\begin{tabular}{@{}l c c c c@{}}
\toprule
Store & Tracked runs & Tier 1 & Tier 2 & Tier 3 \\
\midrule
Quantum-chemistry benchmark (CS1) & 0 & 7 & 0 & 0 \\
Butane MD (CS2) & 6 & 8 & 0 & 0 \\
Modified-DNA MD (CS3) & 116 & 12 & 7 & 0 \\
Docking campaign (CS4), June--July store & 178 & 381 & 85 & 4 \\
Docking campaign (CS4), earlier pilot store & 17 & 3 & 24 & 0 \\
\midrule
Total & 317 & 411 & 116 & 4 \\
\bottomrule
\end{tabular}
\end{table}

\begin{table}[htbp]
\centering
\caption{Project-tier entry inventory by type.}
\label{tab:inventory}
\small
\begin{tabular}{@{}l c p{8.2cm}@{}}
\toprule
Type & Count & Example title \\
\midrule
Scientific finding & 23 & ``P2Rank restraints protocol yields superior scoring over experimental site restraints'' \\
General observation / correction & 40 & ``Mutations are NOT additive: a single-residue variant shifts the binding mode and proves context-dependence'' \\
Failure pattern & 14 & ``cmrest blind docking unreliable: 30+ point variation from input conformation bias'' \\
Error resolution & 16 & ``Template-built peptide PDBs have garbage geometry from hardcoded offsets'' \\
Parameter guidance & 13 & ``Always use max walltime for cpucourt partition (72h)'' \\
Success pattern & 10 & ``N-terminal ALA extension outperforms C-terminal: $+16$ AIR-free vs parent'' \\
\bottomrule
\end{tabular}
\end{table}

The staged pipeline (tool-log capture and conversation distillation writing to \texttt{staging/}, confirmation via \texttt{memory\_confirm} moving entries to \texttt{entries/}) operated continuously across sessions, with the distillation pass triggered at startup and on a 20-minute timer. Confirmation tracks sustained use: the butane campaign (Case Study 2) left eight entries in staging and confirmed none after its short run, whereas the six-week docking campaign contributed the large majority of the confirmed entries above.

\subsection{Temporal Evolution of the Memory Mix\label{subsec:temporal}}

\begin{figure}[htbp]
\centering
\includegraphics[width=\linewidth]{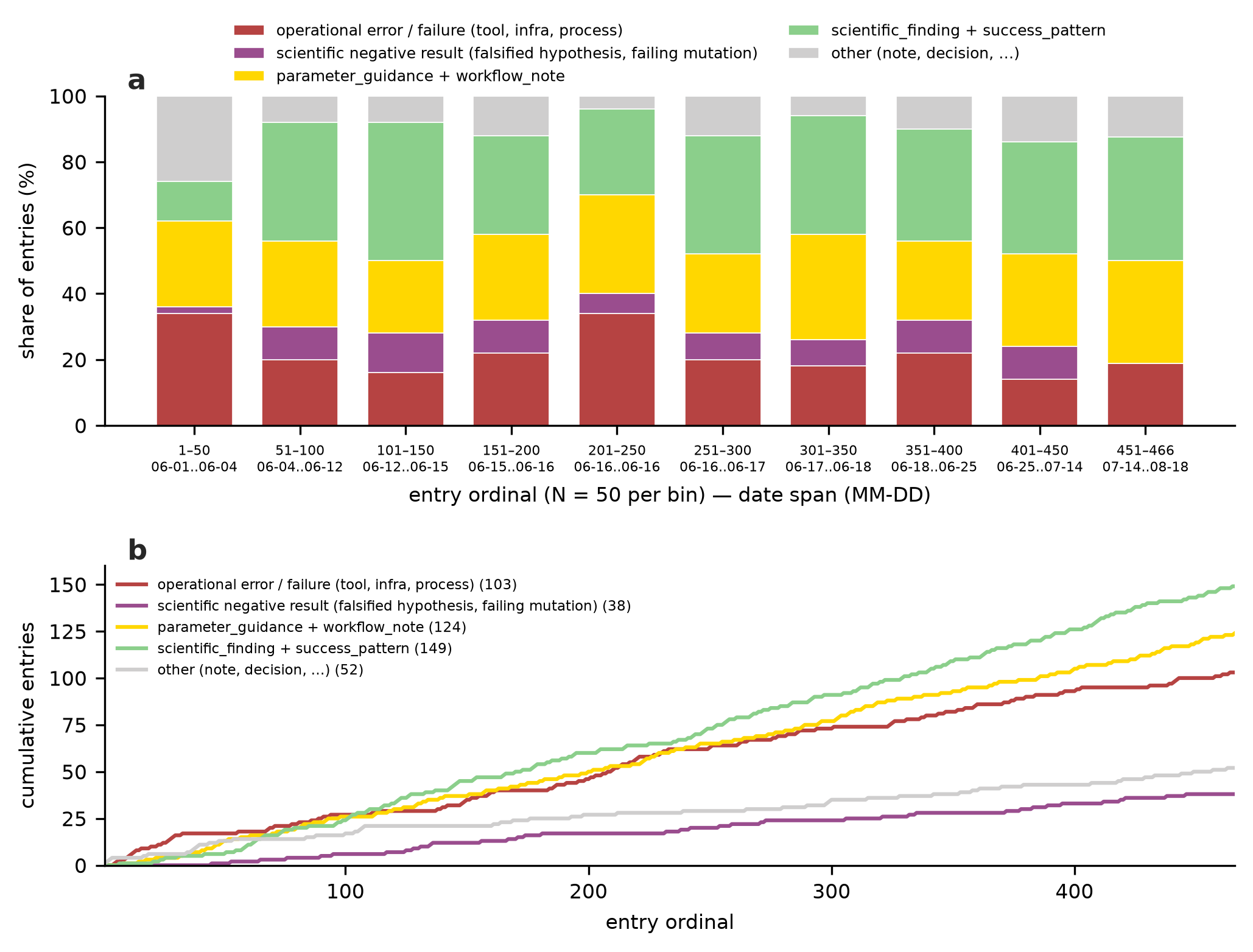}
\caption{Learning-type mix (a) and cumulative entries by learning type (b) across all entries (staging and confirmed) of the docking campaign's June--July store (466 entries at the snapshot of 2026-09-02; Table~\ref{tab:stores}). Entries are plotted in chronological order in bins of 50: nine complete bins span 2026-06-01 to 2026-07-14; a tenth, partial bin holds the 16 remaining records (all staging, dated 2026-07-14 to 2026-08-18), so all 466 records are plotted. Panel (b) gives the same record as cumulative absolute counts on the same entry-ordinal axis. Error and failure entries are split into operational (tool, infrastructure, process) and scientific negative results (falsified hypotheses, failing mutations, method limitations) using keyword rules with manual review of every title, and the operational entries carry the class tags tabulated in ESI Table~\ref{tab:error-classes}. The remaining colour groups combine the other recorded types: parameter guidance with workflow notes, scientific findings with success patterns, and other notes, observations and decisions.}
\label{fig:entrymix}
\end{figure}

The static inventory of Table~\ref{tab:inventory} does not reflect how the composition of the store changed as the project matured. Figure~\ref{fig:entrymix} tracks the learning-type mix across all entries (staging and confirmed) of the docking campaign's June--July store, binned by entry ordinal (N = 50 per bin). The store holds 466 entries at the snapshot of 2026-09-02; the figure shows all of them, spanning 2026-06-01 to 2026-08-18. The same entries are shown as cumulative absolute counts in panel (b). New error classes concentrate early (14 by June 15, 12 during June 16--18, and 2 from June 19 onward; ESI Table~\ref{tab:error-classes}). Error and failure entries were further split into \emph{operational} (tool, infrastructure, process) and \emph{scientific negative results} (falsified hypotheses, failing mutations, method limitations) using keyword rules with manual review of every title. Three features stand out. First, early memory is dominated by operational errors (34\% in bin 1) and parameter guidance; scientific findings and success patterns then rise from 12\% to a sustained 26--42\% share: the store transitions from debugging the workflow to accumulating domain knowledge. Second, the operational-error share does not decline monotonically: it spikes to 34\% on the first mass batch-submission day, run under network instability (rsync failures, duplicate submissions, stale remote outputs; bin 201--250, 2026-06-16), and rises again modestly with the introduction of new contact-analysis and scoring protocols (bin 351--400, 2026-06-18 to 06-25); it falls back after each increase. Third, scientific negative results (falsified hypotheses, failing mutations, method limitations) were captured continuously rather than in isolated episodes (purple, 2--12\% in every complete bin). One dense cluster (bin 251--300, 2026-06-16 to 06-17) records a peptide that failed to converge under the full protocol and the discovery that the simplified protocol's quality metrics were misleading, findings that a conventional workflow would have discarded. We emphasize, finally, that these six weeks cover the exploratory phase of the project, during which the workflow itself changed repeatedly; each protocol change legitimately generates new operational error classes, and a longer observation period will be needed before the composition trends reported here can support stronger statistical conclusions.

How should this trajectory be read? The front-loading of operational errors, together with their reappearance only when a new workflow layer is introduced, is what a burn-in process would produce: each failure mode is recorded once, converted into a rule or a fix, and stops generating new entries. The observations span six weeks of separate sessions, so this silence indicates retention across sessions rather than a quiet period within one. The maturation of the mix, from operational debugging toward scientific findings, mirrors the intended trajectory from procedural friction to domain knowledge. Two caveats bound this reading. First, entry counts measure what was recorded, not what was avoided: the absence of repeated entries for a captured error is consistent with learning but does not demonstrate non-recurrence. The same store documents one clear recurrence failure: a batch of jobs submitted without the by-then-standard memory allocation, one day after that allocation had been established, was lost to out-of-memory errors within minutes. Capture alone does not guarantee retrieval at the decision point. Second, binning by entry ordinal compresses bursts of activity (two bins can span a single day), so the figure reads as a learning-process narrative rather than a uniform time series. A controlled memory-on versus memory-off comparison, which the present case-study design cannot provide, is reported in Section~\ref{subsec:replay}; isolating the contribution of individual tiers remains open.

\subsection{Rules and Promotion}

Keyword tagging of the 103 operational error and failure entries (all tiers) in the docking campaign resolves them into 28 classes (ESI Table~\ref{tab:error-classes} lists all of them, with promotion dates and post-rule recurrences); Table~\ref{tab:inventory} counts only confirmed project-tier entries. Four classes were promoted into durable rules and checklists during the campaign: concurrent-job memory limits (June 1), partition walltime caps (June 5), pre-submission input and restraint validation (June 16), and the per-stage scoring-weight convention (June 18). Promotion was not ranked by frequency: a class was promoted when the mistake concerned knowledge the researcher already held, so the fix was written into a rule at once. Classes with a different natural resolution stayed unpromoted: transfer and rsync failures (the most frequent class, seventeen events) came from connection instability during a heavy batch-submission period and ended with an infrastructure change; the peptide-PDB construction lessons moved into the docking-preparation procedure, after which no peptide-PDB failure was recorded; the memory-tooling class collects distinct defects of the system itself, each repaired once; and BindCraft was set aside by decision, as the tool is not suited to this project. In the recorded store, none of the four appears again after its rule date, while classes that were captured but never promoted continued to be recorded: transfer and rsync failures recurred sixteen times through July 13, the contact-map parsing bug of June 18 was recorded again on June 23, and a stale-output failure was recorded on June 16, after the June 1 practice note. Across the campaign, 19 of the 24 classes without a promoted rule recurred at least once. This contrast is observational and drawn from recorded entries only, as activity regime and infrastructure changes also affect what is recorded; it is consistent with the framework's division of labor (capture preserves a lesson for consultation, an enforced rule changes what the agent checks) without isolating the cause.

These promotions were made by hand, before the proposal pipeline existed; the pipeline now drafts candidates automatically (Methods). What reaches the researcher for periodic review is the consolidation queue (same-claim entries whose titles differ, merged only on approval) and rule-elevation candidates; routine merging and admission are automatic. Because unpromoted entries are still surfaced when relevant (at session start, before a task, and at action time), promotion governs what becomes cross-project doctrine, not whether knowledge is used. Ranking candidates by observed usefulness, and measuring the review time they cost, remain future work.

\subsection{Controlled Replay: Memory On versus Memory Off\label{subsec:replay}}

The store analysis above is observational: it shows what the campaigns deposited, not whether the deposits change what the agent does. To isolate the contribution of memory at the level of single decisions, we replayed four project-internal tasks under controlled conditions. Each task was run with the memory system disabled (no rules, no store, no handover) and with the full agent setup, using the same underlying language models (DeepSeek v4.1-flash and Kimi K2.7), at least two runs per condition and model, with answers graded automatically against keys frozen in advance and derived from the store. A third, oracle arm, in which the decisive facts were pasted directly into the prompt, served as a capability control: where the oracle passes and the memory-off arm fails, the failure reflects missing project knowledge rather than insufficient reasoning ability. Table~\ref{tab:replay} reports the scores.

\begin{table}[htbp]
\centering
\caption{Controlled replay by model (ds = DeepSeek v4.1-flash, kimi = Kimi K2.7; two or more runs per model per condition). Rows 1 and 4 concern HADDOCK3 docking; rows 2 and 3 concern covalent docking with Gnina. Rows 1--2 list each run's score out of the graded total: for example, the two ds memory-off docking-configuration runs scored 1 and 4 of the 10 checks, while all four memory-on runs scored 10 of 10. Rows 3--4 report runs that met the task criterion out of runs attempted, since there the run is the smallest unit that can pass; attempts include runs that stopped without producing an answer (four in the ds flexible-range condition), counted as failures. The oracle runs received the decisive facts in the prompt.}
\label{tab:replay}
\small
\begin{tabular}{@{}l cc cc cc@{}}
\toprule
 & \multicolumn{2}{c}{Memory off} & \multicolumn{2}{c}{Memory on} & \multicolumn{2}{c}{Oracle} \\
\cmidrule(lr){2-3} \cmidrule(lr){4-5} \cmidrule(lr){6-7}
Task & ds & kimi & ds & kimi & ds & kimi \\
\midrule
Docking-configuration recall (10 checks) & 1, 4 & 4, 4 & 10, 10 & 10, 10 & 10, 10 & 10, 10 \\
Covalent-docking protocol recall (13 items) & 1, 1 & 0, 1, 1 & 13, 13 & 12, 12 & 13, 13 & 13, 13 \\
Warhead production conventions (2 runs per cell) & 0/2 & 0/2 & 2/2 & 2/2 & 2/2 & 2/2 \\
Flexible-residue range (1 value per run) & 0/5 & 1/3 & 3/3 & 2/2 & 3/3 & 2/2 \\
\bottomrule
\end{tabular}
\end{table}

Three features of the table deserve comment. First, the memory-off failures are knowledge gaps, not reasoning failures: the oracle arm, given the decisive facts, passed the checks that distinguish the arms in every task. In the warhead-definition task, for example, every memory-off run wrote chemically correct products for the reactive group through which each ligand binds the protein (its warhead, whose post-reaction form the docking program must be given explicitly); what they missed were the campaign's production conventions, the choices no amount of chemical reasoning can supply because they live only in the project's memory. Second, the recurring imperfection in the memory-on arm is traceable to the store rather than the machinery: the only item still missed in the covalent-docking task (both Kimi memory-on runs scored twelve of thirteen, missing only a locked random seed) is absent from the store entry, and an earlier version of the store that contained an outdated protocol flag was faithfully reproduced by memory-on runs (eleven to twelve of thirteen items) until the entry was corrected at the source, after which the task passed in full. Memory propagates what it contains, including its errors; correcting the store repairs downstream behavior without any change to the agent or the model. Third, the flexible-residue range task (the residue-number pair that defines which part of the peptide is allowed to move during refinement) shows a different memory effect: with memory, every run terminated with the correct value (deepseek three of three, Kimi two of two); without it, the deepseek attempts mostly stopped without producing an answer (four of five, and the one that finished was wrong), while Kimi answered correctly once in three. Because one memory-off run derived the required numbering convention on its own, we treat this task as directional evidence only.

These replays are case studies, not a benchmark: four tasks, one project's knowledge, a handful of runs per condition. They do not measure campaign-level time savings, the contribution of individual memory tiers, or performance against external systems. We present them as a controlled complement to the longitudinal analysis above: the store analysis shows what accumulated; the replay shows that what accumulated changes what the agent does.

A hand-written notes file such as AGENT.md would likely work at the scale exercised here (a few dozen lessons fit comfortably in a prompt), but its limits appear as the store grows: the docking campaign's June--July store alone holds more than 600~KB of lessons once entry metadata is stripped (roughly 160{,}000 tokens; about 46{,}000 for the confirmed entries), against an always-loaded boot context below 35~KB, and a flat file carries no confidence scores, recurrence counts, or conflict checks to keep stale or wrong notes from aging like good ones. Long, largely irrelevant contexts are also used less reliably \cite{lostinthemiddle2024,levy2024same} and their compute cost grows faster than linearly with length \cite{tay2022efficient}, which is why the reviewed tier routes knowledge by content class (small always-on rules, on-demand skills) and retrieves project evidence by relevance under a budget rather than wholesale.

\subsection{Operational Measurements and Scale\label{subsec:ops}}

The memory system's operational cost is modest at the scale exercised so far (Table~\ref{tab:ops}). A full session boot, which scans the session logs, merges the rolling handover, and assembles the context, takes tens of seconds to about two minutes depending on the session backlog; retrieving task-relevant memory takes about ten seconds, over ninety percent of it in the optional re-ranking pass of the small model that maintains memory rather than in the local scoring. In the busiest project, that model processed on the order of 1.6 million tokens in one month (about 2,900 calls). No project's store, session logs included, exceeds a few tens of megabytes, and the boot context is below 35~KB. The system has been exercised at the scale of a research group's projects: around one hundred confirmed entries and hundreds of tracked runs per project. The 10,000-run regime has not been tested; storage grows linearly with runs and sessions, while boot and retrieval latency at that scale are not benchmarked. We report the current scale as the evidence boundary rather than extrapolate beyond it. The substrate is deliberately modular: the store is plain Markdown behind an MCP server. A standards-based graph memory with explicit semantic, episodic, and procedural memory classes, provenance, and declarative validation of updates and supersession is an active research direction (the KGmemory project)~\cite{navet-kgmemory-2026}. The current store's provenance is entry-level (source, session identifiers, timestamps) rather than decision-level: it does not record whether an entry reached the project tier by human confirmation or by automatic admission. We intend to adopt it as it matures, carrying the same knowledge with decision-level provenance and declarative conflict handling at larger scale.

Distillation is deliberately eager: the extractor records liberally at session level and the later stages filter, so staging holds provisional material by design. At the 2026-09-02 snapshot, 274 of the docking campaign store's 380 staging entries (72\%) had been observed once only; over the campaign, the consolidation pass merged away 92 duplicate records in 13 runs, and the researcher rejected 5 further merge proposals, which are recorded so they do not resurface. We do not report precision or recall for the extraction pass: no annotated ground truth exists for what a session should yield. The controls in place are the confidence rubric, recurrence- and conflict-gated admission, consolidation, and human confirmation, and using recurrence statistics to tune extraction and admission is under active development.

\begin{table}[htbp]
\centering
\caption{Operational measurements at the scale exercised so far (late September 2026). Boot and retrieval combine the controlled-replay sandbox and the busiest live project. The replay cost is the per-run model spend; the second model used in the replay runs on a subscription plan and is not billed per token.}
\label{tab:ops}
\small
\begin{tabular}{@{}p{0.38\textwidth}p{0.56\textwidth}@{}}
\toprule
Operation & Measurement \\
\midrule
Session boot (scan, handover merge, context) & 34--109 s, dominated by the store scan and the handover merge \\
Memory retrieval (\texttt{memory\_get\_context}) & 8--13 s; over 90\% in two model calls (re-ranking); local scoring 50--700 ms \\
Distillation call & 1--6 s per call (small model) \\
Model volume for memory maintenance & $\sim$1.6 M tokens in one month (about 2{,}900 calls), busiest project \\
Store size, session logs included & below 20 MB (largest project); boot context below 35 KB \\
Replay run cost & \$0.0005--\$0.034 per run, flash-tier API pricing \\
Scale tested & about 100 confirmed entries and hundreds of tracked runs per project; 10{,}000 runs untested \\
\bottomrule
\end{tabular}
\end{table}

\section{Discussion}

Magnolia shows that persistent, tiered memory changes how an AI agent engages with computational chemistry workflows. The shift is not architectural (the tools are the same, the scheduler is the same, the docking software is the same) but operational: the agent that runs a docking protocol on Tuesday knows what it learned on Monday, and that knowledge shapes what it checks, what it flags, and what it presents to the researcher.

Magnolia claims no autonomy: it is best read as a memory and governance layer on top of general-purpose agent architectures, not as a specialized competitor to them. Working with it follows the trajectory of training a colleague: an intern starts on routine tasks under close supervision and requires frequent correction and direction from the researcher; because each correction is captured in the project store, the agent keeps learning and progressively needs less of both. Setup is one conversation, and interaction is in natural language with no command set to learn: what remains for the human is direction and a periodic review of what the agent presents, not continuous supervision.

As the store grows, the agent's contribution shifts from execution to evidence: it surfaces conflicts and cross-run patterns that push the project forward. Case Study 4 is the clearest instance, and the cmrest analysis shows the same shift without any formal rule: cross-protocol comparison of accumulated run histories exposed an input-conformation bias that no single run could show. This is a form of learning carried by the persistent availability of structured evidence rather than by promoted knowledge; whether the cmrest finding generalizes to other protein families remains open, with the recorded entry preserving the evidence and its boundary conditions.


Cumulatively, this changes what the agent is: confirmed entries and promoted rules (human-gated; Section~\ref{sec:memanalysis}) extend what it can be trusted to execute, and routine work begins to draw on retained knowledge rather than rediscovered practice. Competence accumulates in the store rather than in the model: correcting an entry repairs downstream behavior without retraining. Automatic contradiction detection remains future work.

The boundary between what the agent executes alone and what it works through with the researcher therefore moves only as confirmed knowledge accumulates: routine tasks whose outcomes existing tools can verify are delegated; open problems stay collaborative. The first such delegation has been exercised: in an ongoing in-house ligand-design campaign (target withheld), design ran in rounds with each round's outcome confirmed into the project store before the next was planned. A confirmed negative result (tuning the existing contacts of the three best compounds yielded no improvement) redirected the next round toward structural modifications pre-screened by an agent-built accessibility script, which produced two new compounds improving the inherited best score from $-7.7$ to $-11.8$ kcal/mol, with each designed contact verified in the docked pose. This result is preliminary (docking stage; orthogonal validation under way); we cite it as evidence that the training-loop progression is exercisable today, not as a chemical claim.

Two further constraints bound the current work. The compounding effect of memory across whole campaigns remains unmeasured: the case studies cannot isolate how much of the observed change is attributable to memory per se versus the researcher's increasing domain expertise over the same period. The reviewed tier's utility also depends on the generality of what it contains: rules like {\tt ``template-built PDBs have unphysical geometry''} generalize across all peptide docking projects, while rules like ``pocket prediction misses this receptor's shallow groove; derive restraints from crystal contacts instead'' may not. The framework relies on the human promoter to make this judgment; the task-scored retriever (\texttt{memory\_get\_context}) provides a partial safety net, since project-tier entries that are not rules can still be recalled by relevance when the agent encounters a similar context, even in a different project.

Beyond computational chemistry, the memory architecture offers a basis for adaptation to other domains: any computational science where protocol knowledge is scattered across directories, scripts, and lab notebooks faces the same fragmentation problem, and the three-tier memory model, the evidential stance, and the human-gated promotion path are not specific to docking or quantum chemistry. The tool wrappers would change; the memory loop would not. Adaptation would be mostly packaging: create the project and its goal file, register the domain's tools (MCP wrappers only for software driven routinely, command-line wrappers for the rest), and let rules and skills accumulate through use. The wrapper mechanism is agnostic to licensing or distribution, and the harness is replaceable in the same way: the memory and tool servers are external MCP services, so another agent system can host the loop once the harness-specific adapters are supplied, a direction we intend to pursue.

\section{Conclusion}

This paper addressed the fragmentation of protocol knowledge in computational chemistry: the systematic loss of lessons learned between projects and the rediscovery of known errors. Magnolia meets this problem with a 3-tier persistent memory architecture: session observations are automatically captured and distilled, confirmed entries accumulate at the project tier, and human-validated rules govern subsequent behavior across projects.

Across its four case studies, the framework accumulated 116 project-tier entries from 317 tracked runs. Two operational outcomes illustrate the value of the approach: a binding-site correction triggered by a conflict between literature-defined and predicted-pocket restraints, made visible by the agent's cross-run score histories and resolved by its literature search; and a 26-point input-conformation bias in blind docking detected through cross-protocol comparison of accumulated run histories. 

The central insight is that the agent's role is evidential, not prescriptive. Magnolia organizes and surfaces evidence; the human researcher recognizes patterns and makes decisions. Promotion remains the throughput bottleneck: every elevation waits on the researcher, and learnings accumulate faster than they are elevated into durable, cross-project knowledge.

The self-reflex consolidation pass is now implemented for duplicate-merging: an LLM proposes merges, a human confirms them, and a nested Git repository provides reversibility. The immediate next step is routing rule-elevation proposals into the researcher's regular review workflow, so that accumulation stops outpacing generalization. Beyond that, accumulated rules let the agent plan and execute routine experiments, with supervision reserved for open questions. The architecture is tool-agnostic and applies to any computational science domain where protocol knowledge is currently lost between projects.

\bigskip


\section{Acknowledgments}

This work was supported by the French government through the France 2030 investment plan managed by the National Research Agency (ANR), as part of the Initiative of Excellence Université Côte d’Azur under reference number ANR-15-IDEX-01. The authors are grateful to the Université Côte d’Azur’s Center for High-Performance Computing (OPAL infrastructure) for providing resources and support.

We also acknowledge support from Universit\'e C\^ote d'Azur, the Centre National de la Recherche Scientifique (CNRS), the Institut de Chimie de Nice (ICN), and the Interdisciplinary Institute for Artificial Intelligence (3iA) C\^ote d'Azur (ANR-23-IACL-0001). This work was supported by the ANR through the MetaboLinkAI project (ANR-24-CE93-0012-01 and SNSF 10002786). Louis-F\'elix Nothias acknowledges support through a Chaire de Professeur Junior (ANR-22-CPJ2-0048-01).


\section{Data and Code Availability}

\begin{sloppypar}
The Magnolia framework and its two MCP servers are available at \url{https://github.com/HolobiomicsLab/Magnolia-AI} under the MIT license. The complete run records, configuration files, and memory-store entries of the docking campaign (Case Study~4) include unpublished results from the collaborating groups and cannot be released with this preprint; they will be released once that work is published. The framework-level evidence reported here does not depend on the withheld details.
\end{sloppypar}

\bibliographystyle{unsrt} 
\bibliography{citation} 

\newpage

{\LARGE \bf Electronic Supplementary Information}

\section*{Screening of electronic properties along the COMPAS database.}

\begin{figure}[htbp]
\centering
\includegraphics[width=\linewidth]{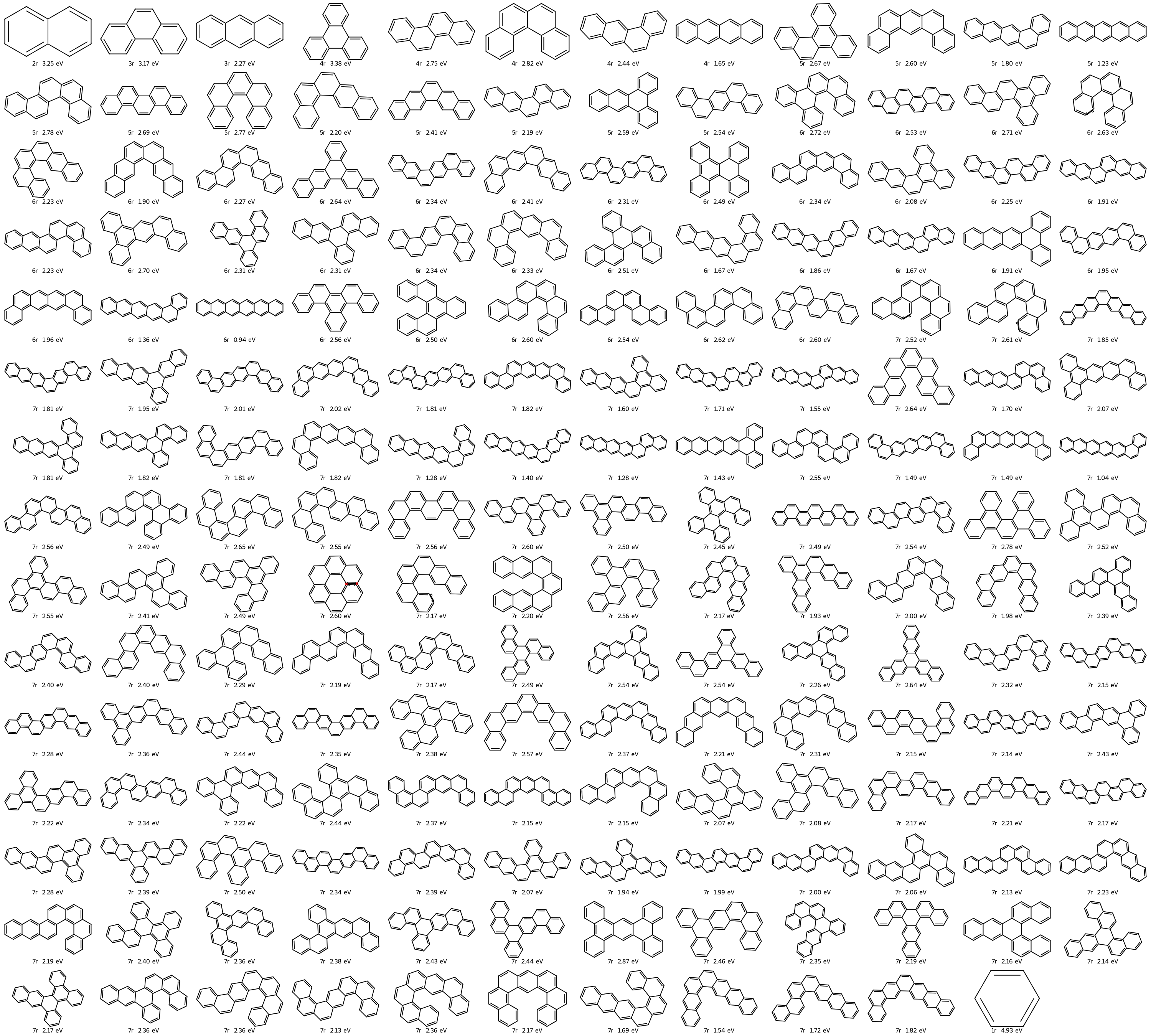}
\caption{Chemical structures of the 179 compounds extracted from the COMPAS database with seven or less rings.}
\label{fig:obp}
\end{figure}


\newpage 
\section*{Molecular dynamics setup of modified oligonucleotides} 

Molecular dynamics (MD) simulations were performed with the Amber2026 suite of programs. The DNA sequences 5$'$-CGCATAT(G/A)-(TA)$_n$-TT-ATGTCAGT-3$'$ were generated using tleap and one non-canonical nucleobase, 5-carboxycytosine or 5-carboxyuracile (see Figure main text). The parameters specific to 5-carboxycytosine and 5-carboxyuracile were obtained using GAFF atom types, with RESP atomic charges computed at the recommended B3LYP/6-31G(d) level of theory. 
Minimizations were performed first with strong restraints on the DNA duplex (500 kcal·mol$^{-1}$·\AA{}$^{-2}$) and then with no restraints for 5000 steps each (2500 of steepest descent and 2500 conjugate gradients). The system was then heated at 300K using a Langevin thermostat. Two series of 1 ns equilibration runs were then performed, first with DNA restraints of 10 kcal·mol$^{-1}$·\AA{}$^{-2}$ and then removing DNA restraints, using a template, which was available in our group. A final production run of 100~ns was then performed for each system. All the trajectories are available at \url{https://doi.org/10.5281/zenodo.23453656}.  

\begin{table}[htbp]
  \centering
  \caption{Composition of the 12 simulated duplexes containing 5-carboxycytosine (XCD) or 5-carboxycytosine (X5D), and characteristics of solvation in truncated octahedral water OPC boxes, with a cut-off of 10\,\AA{}. Seq$_{n}$ differ by the number -- see Figure~5, main text.}
  \label{tab:systems}
  \begin{tabular}{lrrrrrrr} 
    \toprule
    System & {bp} & {atoms} & {OPC waters} & {Na$^+$} & {Cl$^-$}
           & {$V$ / \AA$^3$} & {edge / \AA} \\
    \midrule
    \multicolumn{6}{r}{5-CAC} \\
    \hline 
    XCD seq1 & 20 &  51606 & 12564 & 58 & 20 &  442198 &  95.0 \\
    XCD seq2 & 22 &  73254 & 17943 & 62 & 20 &  616951 & 106.3 \\
    XCD seq3 & 24 &  73010 & 17849 & 66 & 20 &  617316 & 106.3 \\
    XCD seq4 & 26 &  88414 & 21667 & 70 & 20 &  741914 & 113.1 \\
    XCD seq5 & 28 & 111330 & 27363 & 74 & 20 &  922901 & 121.7 \\
    XCD seq6 & 30 & 125190 & 30795 & 78 & 20 & 1036791 & 126.5 \\
    \hline 
    \multicolumn{6}{r}{5-CAU} \\
    \hline 
    X5D seq1 & 20 &  49724 & 12094 & 58 & 20 &  427249 &  93.9 \\
    X5D seq2 & 22 &  59060 & 14395 & 62 & 20 &  505723 &  99.4 \\
    X5D seq3 & 24 &  79056 & 19361 & 66 & 20 &  663611 & 108.9 \\
    X5D seq4 & 26 &  89716 & 21993 & 70 & 20 &  750726 & 113.5 \\
    X5D seq5 & 28 & 114212 & 28084 & 74 & 20 &  944750 & 122.6 \\
    X5D seq6 & 30 & 132440 & 32608 & 78 & 20 & 1095478 & 128.9 \\
    \bottomrule
  \end{tabular}
\end{table}

\begin{table}[htbp]
  \centering
  \caption{Steps constituting our MD protocol (Amber~26 \textsc{sander}; $\Delta t = 2$\,fs; SHAKE was applied on
  bonds featuring hydrogens, \texttt{ntc=2}/\texttt{ntf=2}; 10\,\AA{} cutoff with PME;
  \texttt{iwrap=1}). Restraints are positional on all atoms of the DNA
  (Amber mask \texttt{:1--2N}). Minimization steps were decomposed into 2500 steps of steepest descent followed by 2500 steps of conjugate gradients.}
  \label{tab:protocol}
  {\small
  \setlength{\tabcolsep}{2pt}
  \begin{tabular}{@{}llllll@{}}
    \toprule
    Stage & Ensemble & Length & Restraints
          & Thermostat / Barostat & Output \\
          & & & (kcal\,mol$^{-1}$\,\AA$^{-2}$) & & \\
    \midrule
    \texttt{min1} & --- & 5000 cycles & 500 & --- & every 50 steps \\
    \texttt{min2} & --- & 5000 cycles & none & --- & every 50 steps \\
    \texttt{heat} & NVT & 100\,ps & 10
          & Langevin ($\gamma = 2$\,ps$^{-1}$), & 1\,ps coords, \\
          & (\texttt{ntb=1}) & (50\,000 steps) & & linear 0$\to$300\,K; \texttt{ntx=1}, \texttt{irest=0}
          & 10\,ps restart \\
    \texttt{equil} & NPT & 100\,ps & 2
          & 300\,K, 1\,atm & \\
          & (\texttt{ntb=2}) & & & (\texttt{ntp=1}, \texttt{taup=2}\,ps); \texttt{ntx=5} & \\
    \texttt{equil\_nr} & NPT & 1\,ns & none (\texttt{ntr=0})
          & as \texttt{equil}, started from & 10\,ps coords \\
          & & (500\,000 steps) & & \texttt{equil.rst7} & (100 frames) \\
    \bottomrule
  \end{tabular}}
\end{table}

\section*{Example of a distilled staging entry}

\noindent The listing below is one complete staging entry, verbatim from a
project store (internal identifiers removed, long lines wrapped to fit the page); it is the parameter-guidance
entry discussed in Case Study~1, whose second observation records the
correction of the initial pre-optimization guidance.

{\small
\begin{verbatim}
---
confidence: 0.8
created: '2026-06-23T10:16:26.894435+02:00'
description: For planar, conjugated polybenzenoid hydrocarbons, adding a GFN-FF
  pre-optimization step before GFN2-xTB is unnecessary because the molecules
  lack conformational complexity. OpenBabel --gen3d followed
observation_count: 2
source: opencode_distill
tags:
- GFN-FF
- pre-optimization
- polybenzenoid hydrocarbons
- planar PAHs
- COMPAS
title: No GFN-FF pre-optimization needed for planar polybenzenoid hydrocarbons prior
  to GFN2-xTB optimization
tools:
- xtb
- OpenBabel
type: parameter_guidance
updated: '2026-06-23T08:36:39.435949+00:00'
---

For planar, conjugated polybenzenoid hydrocarbons, adding a GFN-FF pre-optimization
  step before GFN2-xTB is unnecessary because the molecules lack conformational
  complexity. OpenBabel --gen3d followed directly by GFN2-xTB --opt produced clean
  geometries in all 25 test cases. Caveat: this guidance applies only to rigid,
  planar PAHs; molecules with flexible side chains, multiple rotatable bonds, or
  strained non-planar frameworks might benefit from GFN-FF pre-optimization.

## Observation 2 (2026-06-23)

The COMPAS-2 paper's production code (`conf_gen.py`) uses RDKit ETKDG embedding
  followed by UFF optimization (maxIters=1000) before GFN2-xTB geometry optimization.
  The older `func.py` tested both `uff_optimize` and `mmff_optimize`, but the
  production pipeline settled on UFF only. This is the validated pipeline for COMPAS
  molecules (both pure hydrocarbons and heterocycles). The user's earlier session
  (2026-06-20) used OpenBabel `--gen3d` (which internally uses MMFF94), which does
  not match the paper's approach. The earlier recommendation of GFN-FF
  pre-optimization was incorrect; the correct pre-xTB step is RDKit UFF optimization.
\end{verbatim}
}

\section*{Error-class recurrence in the docking campaign store}

\noindent Each error or failure entry in the docking campaign store was assigned to one of 28 classes by keyword rules with manual review; the table lists each class with its recorded events, first and last dates, promotion status, and post-rule recurrences.

\begin{table}[htbp]
\centering
\caption{Operational error and failure classes in the docking campaign store (all tiers; snapshot 2026-09-02). The 103 entries resolve into 28 classes by keyword rules with manual review. ``Entries'' counts recorded events; ``first'' and ``last'' are the dates (year 2026) of the first and last recorded event. Rule dates mark the four promoted classes; ``after rule'' counts events strictly after the promotion date (0 for all four; events on the rule date itself are not counted there). Single-entry classes never recurred.}
\label{tab:error-classes}
\small
\begin{tabular}{@{}l r l l l r@{}}
\toprule
Class & Entries & First & Last & Promoted rule & After rule \\
\midrule
cluster-trust & 4 & 06-01 & 06-18 & -- & -- \\
oom & 4 & 06-01 & 06-01 & 06-01 & 0 \\
output-collision & 6 & 06-01 & 06-16 & -- & -- \\
p2rank & 2 & 06-01 & 06-01 & -- & -- \\
tooling & 6 & 06-01 & 06-17 & -- & -- \\
transfer & 17 & 06-01 & 07-13 & -- & -- \\
config-invalid & 1 & 06-05 & 06-05 & 06-16 & 0 \\
restart-clean & 5 & 06-05 & 07-13 & -- & -- \\
walltime & 1 & 06-05 & 06-05 & 06-05 & 0 \\
bindcraft & 8 & 06-10 & 06-16 & -- & -- \\
pdb-build & 8 & 06-11 & 06-16 & -- & -- \\
peptide-labels & 2 & 06-11 & 06-16 & -- & -- \\
sbatch-quoting & 2 & 06-12 & 06-15 & -- & -- \\
scoring-weights & 3 & 06-12 & 06-18 & 06-18 & 0 \\
communication & 1 & 06-16 & 06-16 & -- & -- \\
hallucination & 1 & 06-16 & 06-16 & -- & -- \\
memory-meta & 11 & 06-16 & 08-06 & -- & -- \\
process & 1 & 06-16 & 06-16 & -- & -- \\
sampling & 3 & 06-16 & 06-25 & -- & -- \\
submission-hygiene & 3 & 06-16 & 06-16 & -- & -- \\
viz & 2 & 06-16 & 07-24 & -- & -- \\
perspacite & 2 & 06-17 & 06-24 & -- & -- \\
queue & 2 & 06-17 & 06-19 & -- & -- \\
report-convergence & 2 & 06-17 & 07-06 & -- & -- \\
capri-clt & 1 & 06-18 & 06-18 & -- & -- \\
contactmap-parsing & 2 & 06-18 & 06-23 & -- & -- \\
deadlock & 2 & 06-25 & 06-25 & -- & -- \\
misc & 1 & 08-18 & 08-18 & -- & -- \\
\bottomrule
\end{tabular}
\end{table}

\end{document}